\documentclass[journal]{IEEEtran}

\usepackage{graphicx}
\usepackage{subcaption}
\usepackage{amsmath}
\usepackage{booktabs}
\usepackage{siunitx}
\usepackage{threeparttable}
\usepackage{csquotes}
\usepackage{bm}
\ifCLASSINFOpdf
\else
\fi
\usepackage{xcolor}

\begin{document}
%
\title{Off-Axis Compliant RCM Joint with Near-Isotropic Stiffness and Minimal Parasitic Error}
%
%
%

\author{%
Federico~Mariano\textsuperscript{1,2},
Elena~De~Momi\textsuperscript{2},
Giovanni~Berselli\textsuperscript{1,3},
Jovana~Jovanova\textsuperscript{4},
Just~L.~Herder\textsuperscript{5},
and~Leonardo~S.~Mattos\textsuperscript{1}
\thanks{\textsuperscript{1} Biomedical Robotics Lab, Department of Advanced Robotics, Istituto Italiano di Tecnologia, Genoa, Italy.}%
\thanks{\textsuperscript{2} Department of Electronics, Information and Bioengineering, Politecnico di Milano, Milan, Italy}%
\thanks{\textsuperscript{3}DIME, Università di Genova, Genoa, Italy}%
\thanks{\textsuperscript{4}Department of Maritime and Transport Technology , Delft University of Technology, Delft, The Netherlands.}%
\thanks{\textsuperscript{5}Department of Precision and Microsystem Engineering, Delft University of Technology, Delft, The Netherlands.}%
\thanks{Corresponding author: Federico Mariano, \\
\texttt{federico.mariano@iit.it}, \\
\texttt{federicomariano.mariano@polimi.it}.}%
\thanks{This manuscript has been submitted to the IEEE for possible publication. Copyright may be transferred without notice, after which this version may no longer be accessible.}%
}

\maketitle

\begin{abstract}
This paper presents an off-axis, monolithic compliant Remote Center of Motion (RCM) joint for neuroendoscopic manipulation, combining near-isotropic stiffness with minimal parasitic motion. Based on the Tetra II concept, the end-effector is placed outside the tetrahedral flexure to improve line of sight, facilitate sterilization, and allow rapid tool release. Design proceeds in two stages: mobility panels are sized with a compliance-based isotropy objective, then constraining panels are synthesized through finite-element feasibility exploration to trade stiffness isotropy against RCM drift. The joint is modeled with beam elements and validated via detailed finite-element analyses, including fatigue-bounded stress constraints. A PA12 prototype is fabricated by selective laser sintering and characterized on a benchtop: a 2 N radial load is applied at the end-effector while a 6-DOF electromagnetic sensor records pose. The selected configuration produces a stiffness-ellipse principal axis ratio (PAR) of 1.37 and a parasitic-to-useful rotation ratio (PRR) of $0.63\%$. Under a $4.5^\circ$ commanded rotation, the predicted RCM drift remains sub-millimetric ($0.015–0.172$ mm). Fatigue analysis predicts a usable rotational workspace of $12.1^\circ–34.4^\circ$ depending on direction. Experiments reproduce the simulated directional stiffness trend with typical deviations of $6–30\%$, demonstrating a compact, fabrication-ready RCM module for constrained surgical access. 
\end{abstract}

\begin{IEEEkeywords}
Remote Center of Motion (RCM), compliant mechanisms, neuroendoscopy, isotropic stiffness, and additive manufacturing.
\end{IEEEkeywords}

%
\IEEEpeerreviewmaketitle

\section{Introduction}

\IEEEPARstart{S}urgical Robots for Minimally Invasive Surgery (MIS) enforce the Remote Center of Motion (RCM) constraint, allowing the robot to reorient the endoscope while keeping a point along the shaft stationary during the operation \cite{sadeghian2019constrained, zhou2018new, zhao2025compact}. The RCM point lies at the \enquote{burr-hole}, allowing surgeons to reach multiple targets with minimal risk of damaging surrounding tissue.

\begin{figure}[t]
    \centering
    \begin{subfigure}[b]{0.22\textwidth}
        \centering
        \includegraphics[width=\linewidth]{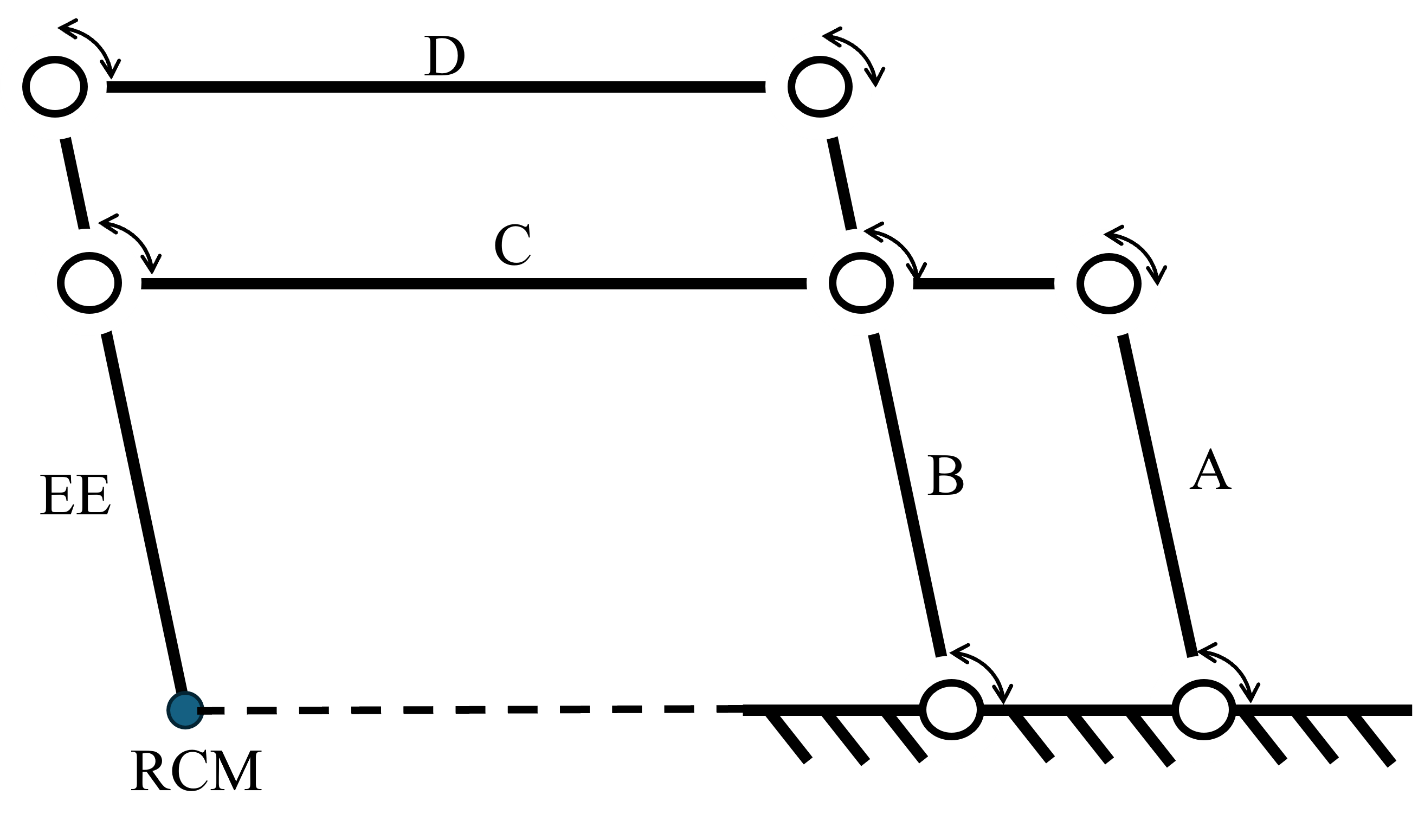}
        \caption{}
        \label{solution1}
    \end{subfigure}
    \hfill
    \begin{subfigure}[b]{0.22\textwidth}
        \centering
        \includegraphics[width=\linewidth]{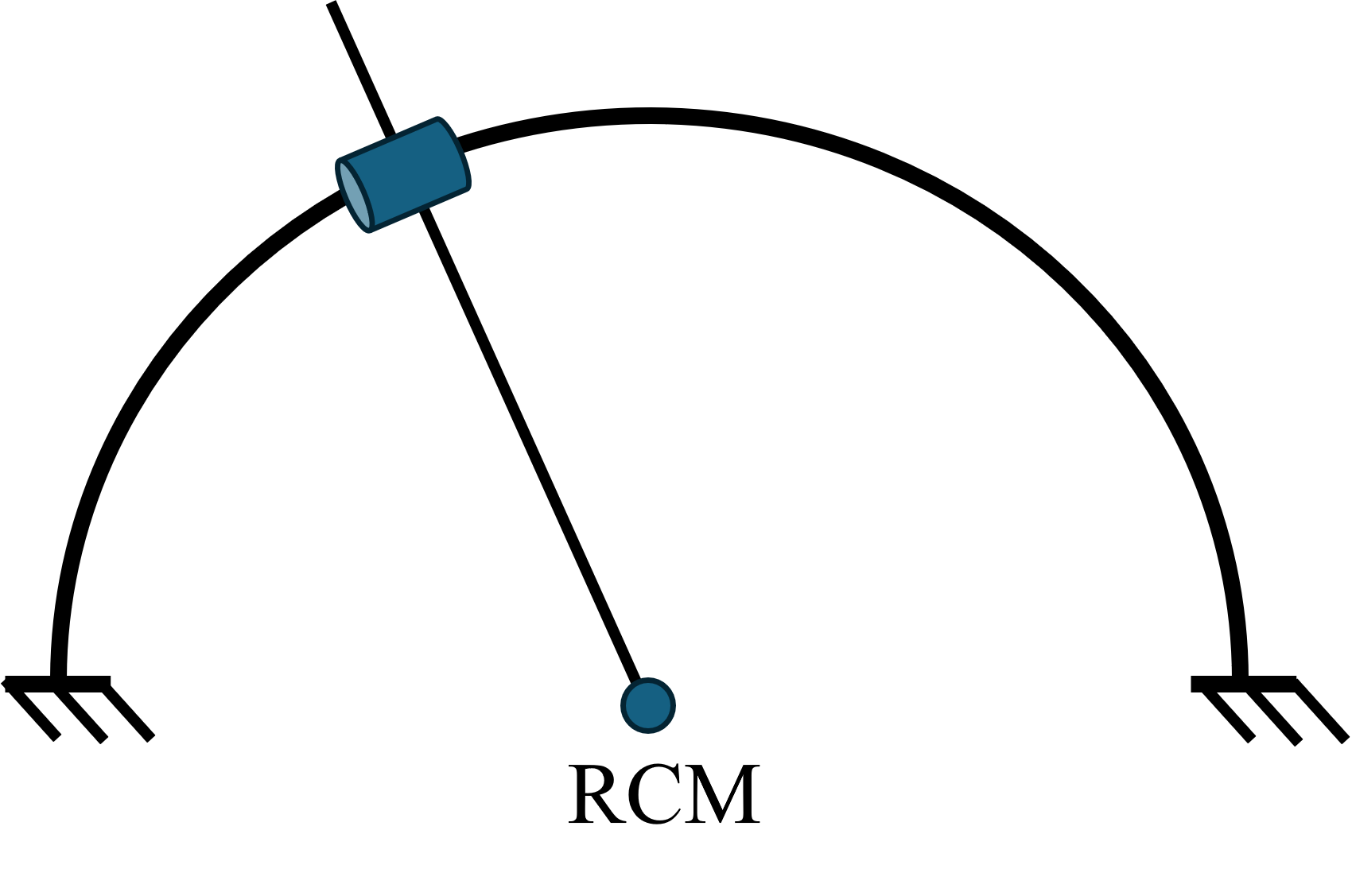}
        \caption{}
        \label{solution2}
    \end{subfigure}
    \caption{\textbf{(a)} Representation of a double parallelogram RCM mechanism. The mutual parallelism of A, B, C, and D forces the prolongation of the EE to pass through the RCM - \textbf{(b)} Representation of a RCM mechanism with circular guides.}
    \label{solution}
\end{figure}

In the literature different mechanisms are described which can be used to enforce the RCM constraint:
\begin{itemize}
    \item \textbf{Double-parallelogram mechanism} (Fig.~\ref{solution1}): Two coupled four-bar loops with moving links A, B, C, and D, plus the End-Effector (EE). Mutual parallelism forces the prolongation of the EE to pass through a fixed point, the RCM \cite{singh2023sensitivity, chen2020novel}.
    This type of mechanisms provides a passive kinematic constraint of the remote center of motion, but they typically rely on multiple classical joints and linkages, which can increase footprint and introduce backlash/friction and tolerance-sensitive accuracy.
    \item \textbf{Spherical mechanism} (Fig.~\ref{solution2}): A set of revolute joints intersecting at a single point that realizes the RCM \cite{li2024design, essomba2018kinematic}.
    Spherical RCM mechanisms can generate rotations about a fixed point by construction, but practical implementations may face limited workspace and their accuracy is sensitive to assembly tolerances and joint clearances.
    \item \textbf{Software RCM}: A general-purpose robotic arm generates motions, while software enforces the RCM \cite{laribi2012design, sadeghian2019constrained, nasiri2024teleoperation}.
    Software-enforced RCM approaches offer high flexibility and can be implemented on general-purpose robotic arms, but the RCM constraint is not passive and depends on calibration, sensing, and control performance. Joint backlash and modeling errors can lead to residual RCM drift.
    \item \textbf{Compliant mechanisms}: Although less widespread, several works have explored compliant joints for biomedical applications \cite{chandrasekaran2018realization, sun2021cruciate}. These mechanisms exhibit the elastic deformation of a monolithic structure: since no discrete parts are assembled through pin or bearing interfaces, the clearance of the joint is eliminated, and the absence of sliding contact greatly reduces friction and wear, often resulting in improved motion smoothness and accuracy compared to traditional architectures. \cite{thomas2021surgical, yang2017design, kabganian2024towards}.
\end{itemize}

\begin{figure}
    \centering
    \includegraphics[width=1\linewidth]{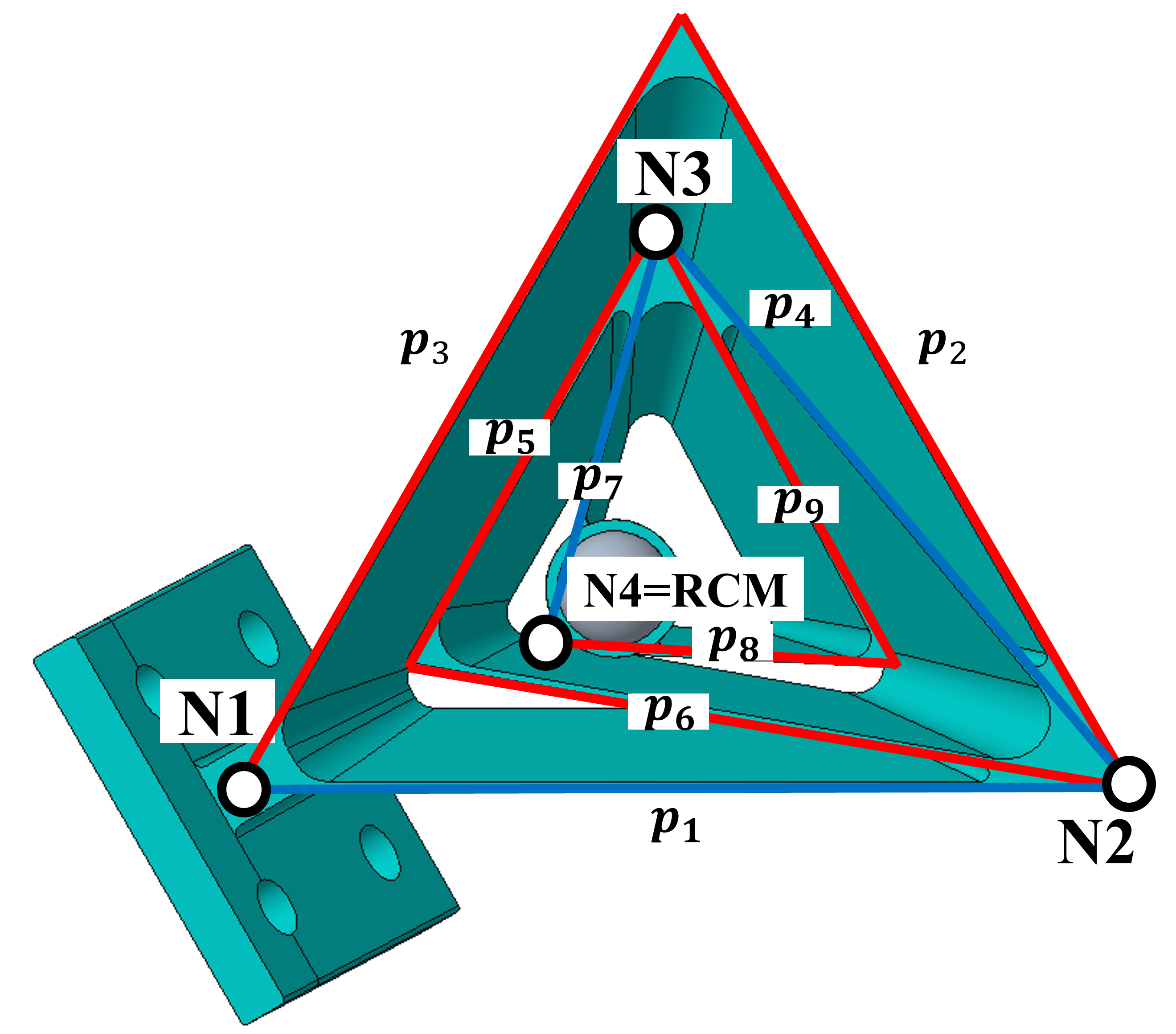}
    \caption{Isometric view of Tetra II joint composed of nine flexural walls ($p_1, ..., p_9$). The blue panels, obtained connected the nodes $N1, N2, N3$ and $N4$, are responsible for joint motion and are therefore the ones to be modified to have isotropic stiffness behavior. The red panels block the joint’s rotation, constraining the tool from following circular rotations around the RCM.}
    \label{fig:walls_tetraII}
\end{figure}

A representative compliant solution is the Tetra~II joint \cite{rommers2021new}, a monolithic compliant joint that reorients its EE while maintaining a fixed RCM. As illustrated in Fig.~\ref{fig:walls_tetraII}, the joint comprises three tetrahedral structures connected in series in a nested configuration. The EE is housed within the structure while the RCM remains fixed.

\begin{figure}
    \centering
    \includegraphics[width=0.8\linewidth]{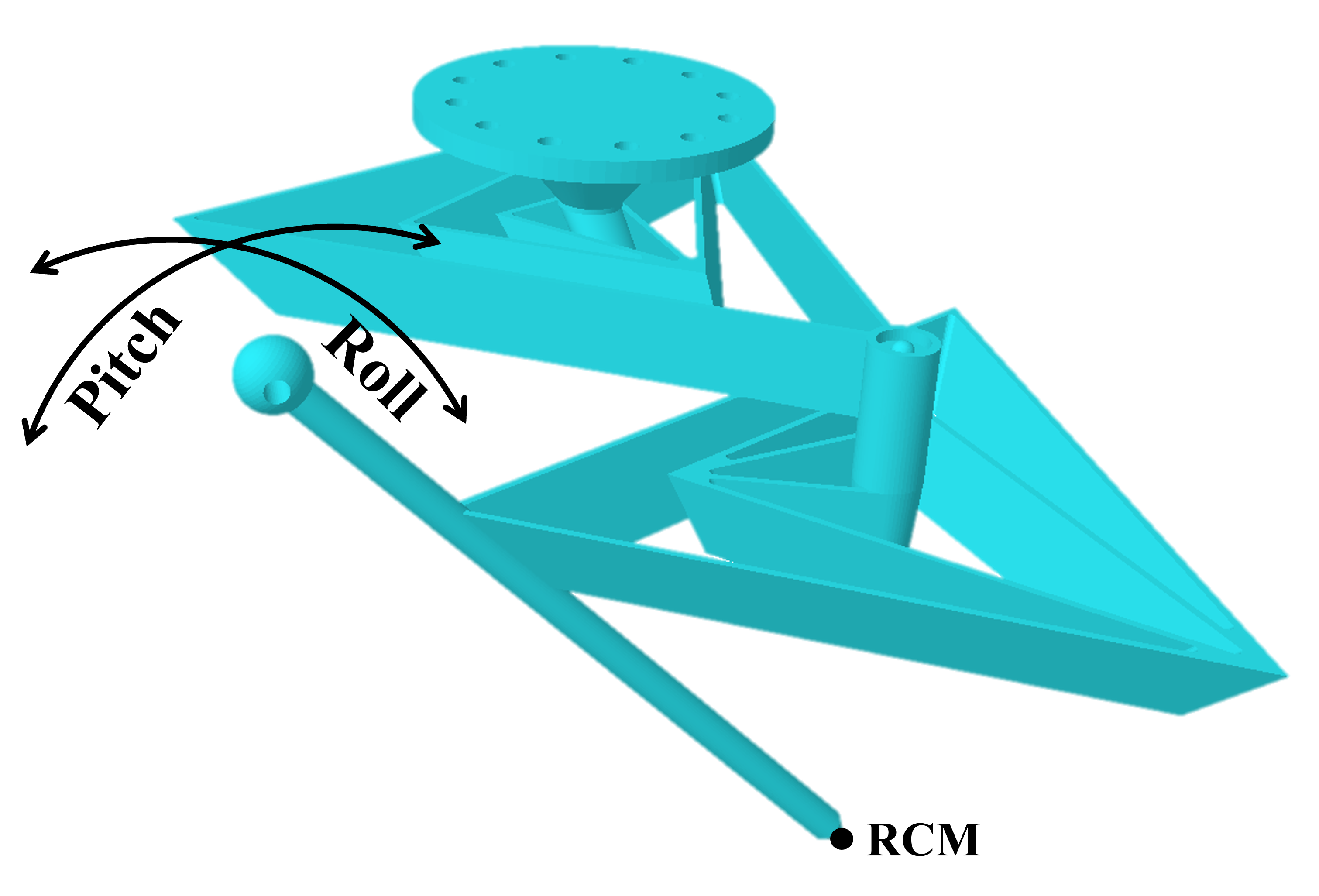}
    \caption{View of dual compliant joint composed of two identical joint connected in series. This joint can move the EE around the RCM along the pitch and roll directions. \cite{mariano2026dual}}
    \label{fig:dual_compliant_joint}
\end{figure}

A first modification of the Tetra II joint was presented through a FEM-based fatigue optimization of the monolithic geometry, to maximize the lifetime of the mechanism while keeping the RCM displacement error below the 0.5 mm limit over the required workspace \cite{mariano2025new}. Subsequently, an off-axis variant, referred to as a dual-compliant joint, was introduced, in which the EE is relocated from the center to the side of the tetrahedral structure (Fig. \ref{fig:dual_compliant_joint}) \cite{mariano2026dual}. This off-axis relocation improves the surgeon’s field of view over the surgical site, simplifies sterilization by functionally separating sterilizable and non-sterilizable parts, and enables rapid removal or exchange of the EE in case of malfunction. However, translating the EE away from the joint center can negatively affect the mechanical behavior of a single compliant RCM stage, typically degrading stiffness isotropy and increasing parasitic RCM shift. The dual-compliant architecture mitigates these effects by adopting a two-stage serial arrangement, allowing the off-axis clinical advantages to be retained while recovering near-isotropic stiffness and reducing the parasitic error at the RCM \cite{mariano2026dual}.

The aim of this paper is to propose a new compliant RCM mechanism that achieves near-isotropic stiffness and low parasitic error without the need to connect two joints in series. We present the design rationale and geometry synthesis, develop analytical and FEM models to predict the stiffness and fidelity to the RCM, and fabricate a monolithic prototype. We experimentally validate the mechanism on a benchtop: a mass that generates a force of 2N is attached at the EE to deflect the structure. At the same time, an electromagnetic tracker measures the motion of the tools field to quantify the displacement across the workspace. The results demonstrate that the proposed off-axis architecture can preserve RCM accuracy while providing an isotropic profile suitable for MIS manipulation.

The main contributions of this work are:
\begin{itemize}
    \item A novel \emph{off-axis}, monolithic compliant RCM joint expressly engineered for future operating room use, achieving near-isotropic stiffness and very low parasitic error in the burr-hole.
    \item A two-stage computational design workflow that (i) employs MATLAB (Mathworks, Inc.) to identify the placement of internal panels to equalize the directional stiffness and (ii) uses ANSYS Workbench 2023 R1 (Ansys, Inc) FEM analysis to search for the configuration that minimizes the parasitic error in the RCM.
    \item An integrated modeling framework that combines analytical formulations and FEM-based analysis to predict the stiffness distribution and fidelity of the RCM across the workspace.
\end{itemize}

\section{Design Requirements} \label{Requirements}
The compliant joint must satisfy the following requirements to ensure safe use in the surgical theater in neuroendoscopy.

\subsection{Safety Requirements}
The requirements listed below must be satisfied to ensure an adequate level of safety for surgical use.
\begin{itemize}
    \item \textbf{Limit on parasitic motion:} Any unintended shift of the RCM under actuation shall not exceed $1\mathrm{mm}$, thus minimizing avoidable trauma to cerebral tissue and remaining consistent with the accuracy of state-of-the-art neuroendoscopic guidance devices \cite{wilson2010evaluating}.
    \item \textbf{Rapid release of the end effector (EE):} The distal tool must be detachable quickly and intuitively, enabling fast instrument exchange and immediate manual override in the event of a malfunction.
\end{itemize}

\subsection{Clinical Requirements}
These requirements must be respected to ensure usability in the operating room.

\begin{figure}[t]
    \centering
    \begin{subfigure}[b]{0.24\textwidth}
        \centering
        \includegraphics[width=\linewidth]{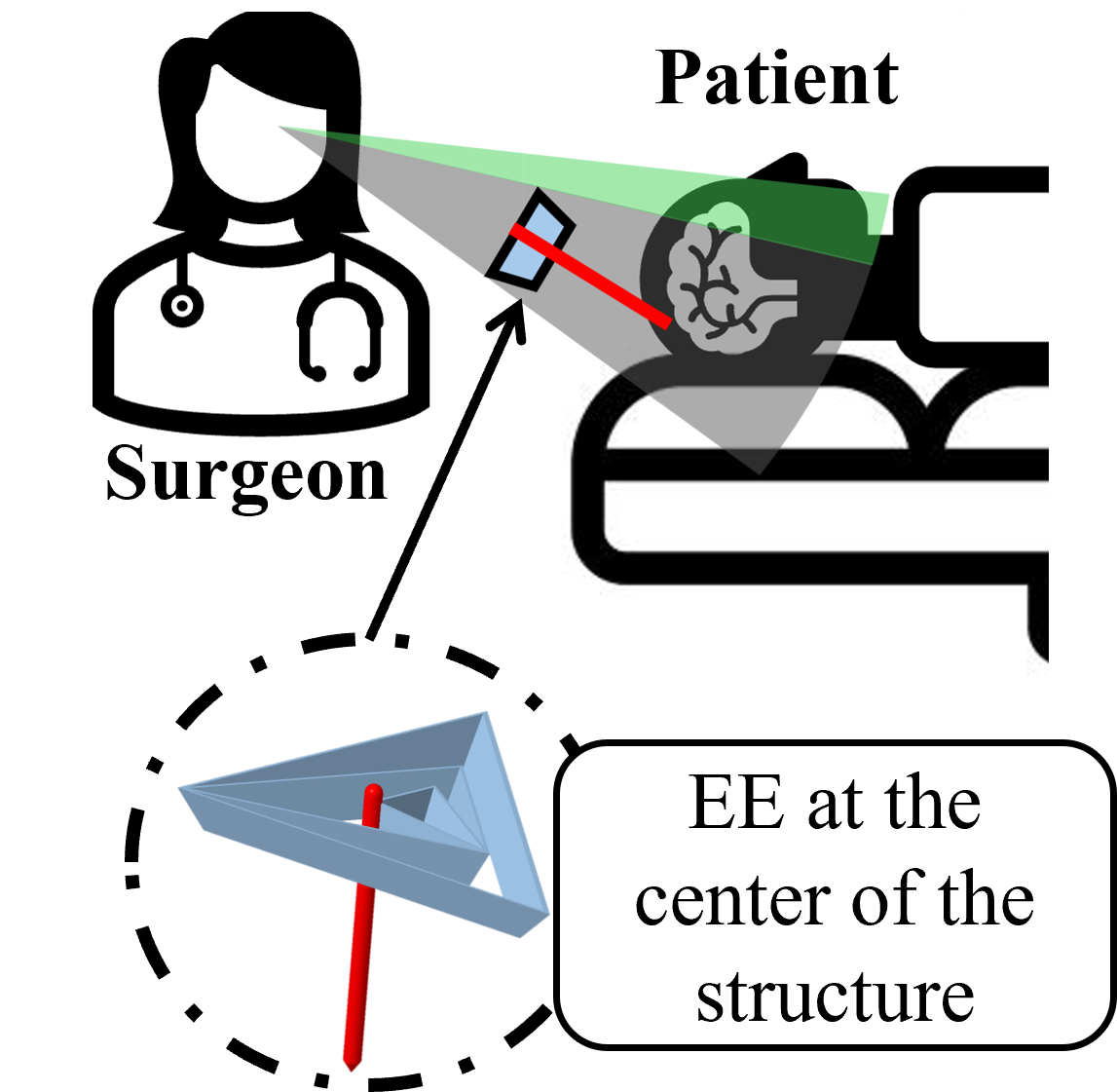}
        \caption{}
        \label{solution3}
    \end{subfigure}
    \hfill
    \begin{subfigure}[b]{0.24\textwidth}
        \centering
        \includegraphics[width=\linewidth]{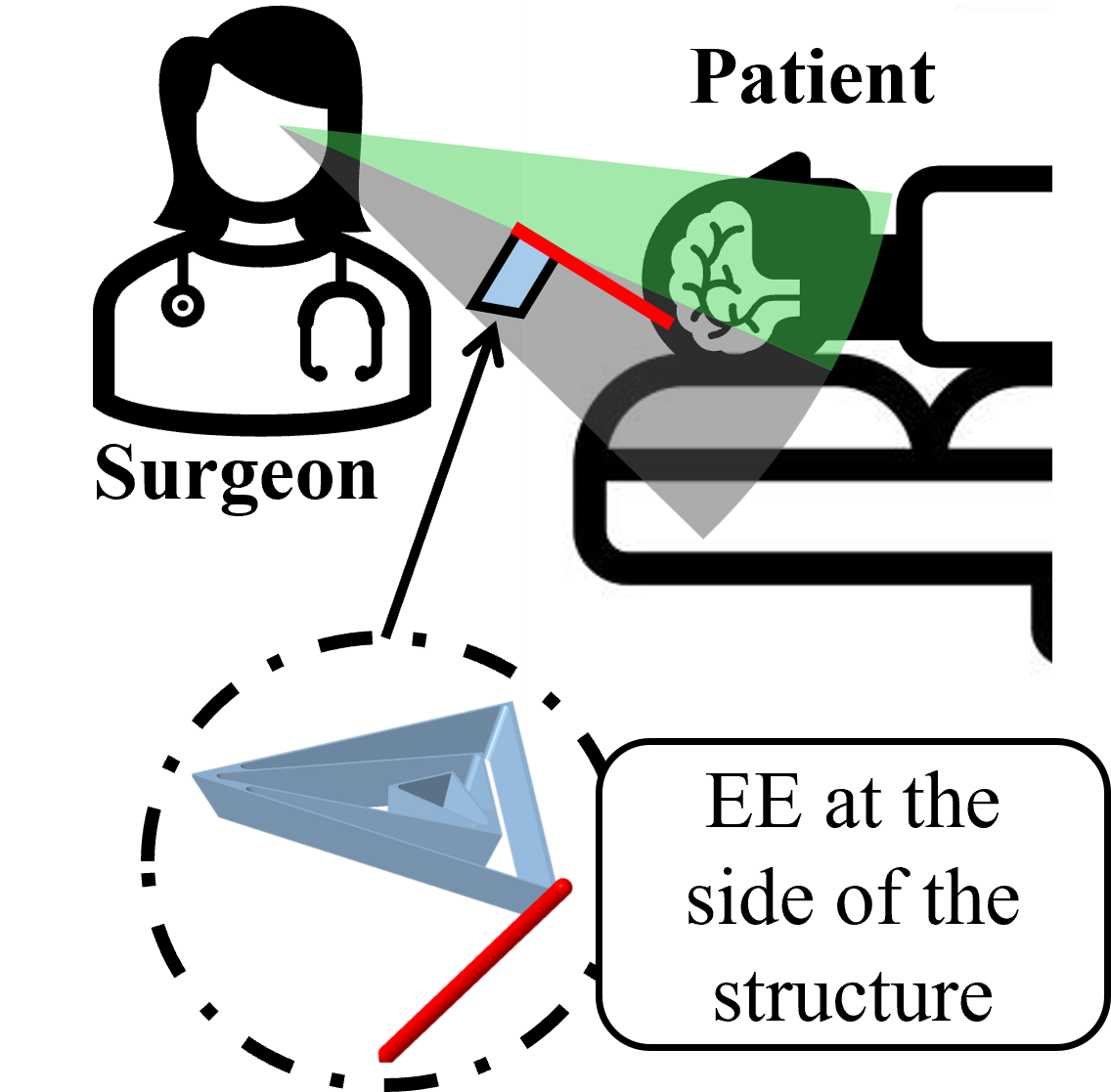}
        \caption{}
        \label{solution4}
    \end{subfigure}
    \caption{Scheme of the line of sight of the surgeon over the entry area. The green cone represents the visible area while the black cone represents the non visible area. In \textbf{(a)} the joint has the EE located at the center of tetrahedral structure, while in \textbf{(b)} the joint has the EE located at the side of tetrahedral structure. The EE is represented with the red line.}
    \label{fig:line_of_sight}
\end{figure}

\begin{itemize}
    \item \textbf{Material constraints:} All elements in contact with the patient or within the sterile field must be manufactured using bio-compatible materials. Any component not covered with a sterile drape must withstand a steam autoclave or an equally effective sterilization method \cite{george2024day}.
    \item \textbf{Unobstructed surgical view:} The mechanism must be shaped and dimensioned to not occlude the surgeon’s line of sight at or near the endoscope entry portal, guaranteeing unobstructed visual access to the surgical field as represented in Fig. \ref{fig:line_of_sight}. 
    \item \textbf{Management of sterility:} The joint must be conceived so that the components that can withstand sterilization are physically separated from those that cannot. Designers must accommodate a sterile drape or equivalent barrier to isolate all non-sterilizable parts.
\end{itemize}

\subsection{Control/Performance Requirements}
These requirements are necessary to enable smooth manipulation and effective control during use.

\begin{itemize}
    \item \textbf{Uniform stiffness behavior:} Around the RCM, the joint must oppose the pitch and roll displacements, shown in Fig. \ref{fig:dual_compliant_joint}, with essentially the same rigidity. Such near-isotropic stiffness is essential for future closed-loop motor control, facilitating future control implementation. Pronounced directional stiffness disparities would impede smooth rotation and degrade performance. It was also noted that the RCM’s parasitic motion tends to be larger along the direction of greater stiffness. Achieving perfectly isotropic stiffness should therefore reduce the parasitic error.

    \item \textbf{Operational workspace.} The literature suggests that only $\sim\!4.5^{\circ}$ is typically needed to contact the ventricular side panels \cite{duffner2003anatomy}. Although we specify a larger angular workspace of $\pm 15^{\circ}$ about the RCM. This margin is intentional: it (i) guarantees overlap with the preplanned trajectory envelope despite setup and registration errors, (ii) accommodates patient/anatomical variability (burr-hole location, ventricle size/shift), and (iii) preserves maneuverability for contingency motions (e.g., inspection, reorientation, or instrument exchange).
\end{itemize}

\section{Materials and Methods}

\subsection{Analytical Identification of the Isotropic Stiffness Criterion}
\label{isotropic}

The \emph{Tetra II} joint comprises nine flexural panels, as shown in Fig. \ref{fig:walls_tetraII}. The flexible pivots correspond to the connection edges between the two tetrahedra, and the intersection of these edges defines the RCM. Three “blue” panels provide the end-effector (EE) mobility, while the remaining “red” panels, defined as torsion stiffeners, constrain rotation about the RCM. Without them, the mechanism would not exhibit a stable and well-defined RCM.
The EE corresponds to a cylinder, which simulates the presence of the endoscope. 

To impose direction-independent rigidity, we model the stiffness contribution of the three blue panels as equivalent slender beam elements under small deflections. This reduced-order formulation yields closed-form stiffness terms for fast isotropy-oriented synthesis, while final geometries are subsequently verified using 3D FEM.

\begin{figure}
    \centering
    \includegraphics[width=0.7\linewidth]{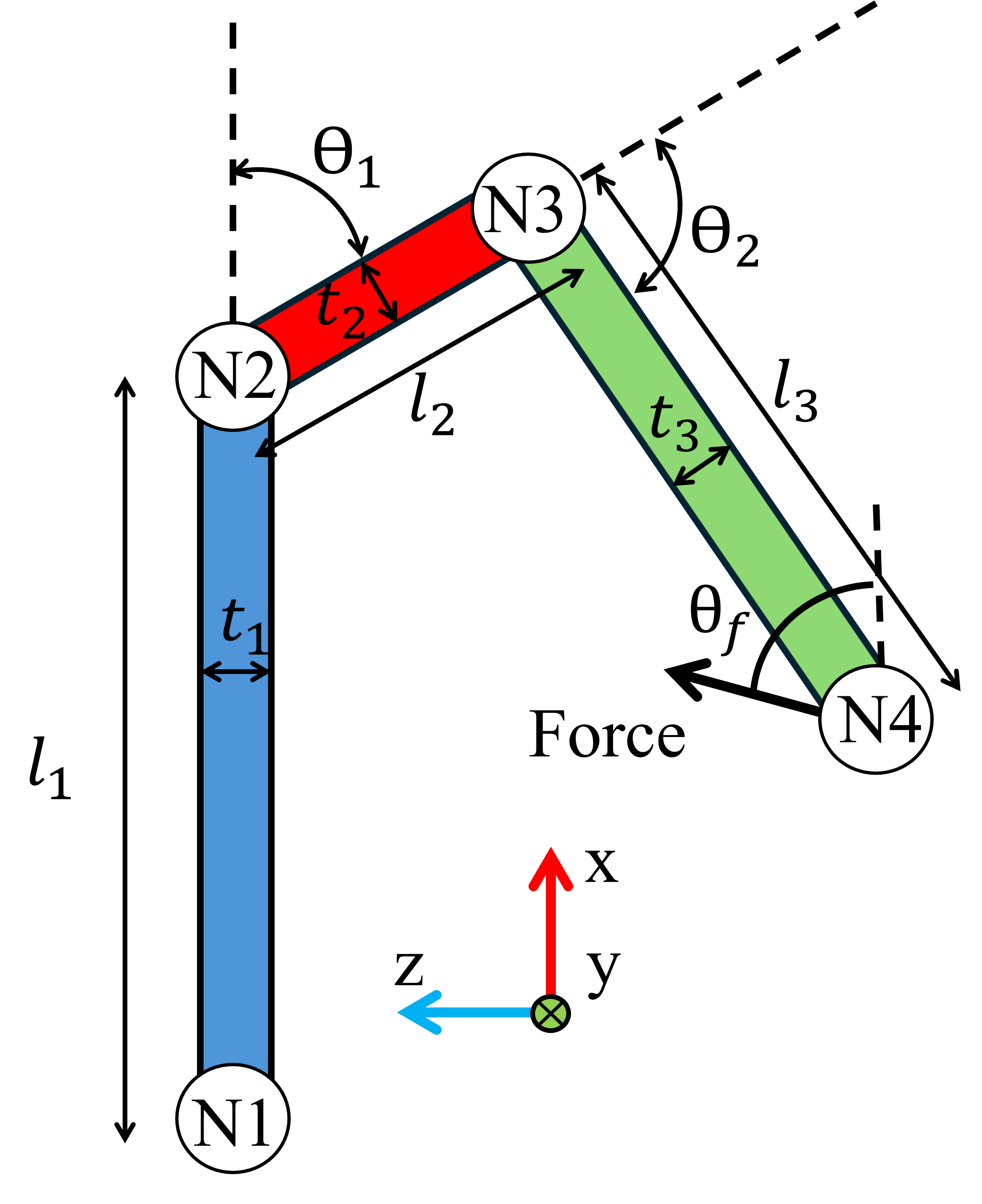}
    \caption{Scheme of the three panels responsible for the joint movement. Highlighted are nodes $N1, N2, N3$ and $N4$, along with the outputs of the optimization $L_1, L_2, L_3, t_1, t_2, t_3, \theta_1$ and $\theta_2$. }
    \label{fig:scheme_tretravi}
\end{figure}

\paragraph{Design variables and assumptions}
We performed the analysis and optimization in \textsc{Matlab}~R2023a. The design vector includes the three panel lengths $\{L_1,L_2,L_3\}$, the three panel thicknesses $\{t_1,t_2,t_3\}$, and the orientation angles between panels $\{\theta_1,\theta_2\}$ (Fig. \ref{fig:scheme_tretravi}). The three \enquote{blue} panels are assumed to be mutually parallel for tractability and not directed toward the RCM. The requirements introduced in Section \ref{Requirements} directly inform the choice of the initial values used in the optimization. In order to obtain a system that does not obstruct the surgeon’s line of sight over the entry site, allows rapid release of the EE, and can be easily covered with a sterile drape, the EE should be located outside the tetrahedral structure rather than at its center, as shown in Fig.~\ref{fig:line_of_sight}.
\\
At this stage, these requirements are reflected in the initialization of the optimization script. The script asks the user to provide starting values for $L_1, L_2, L_3, t_1, t_2, t_3, \theta_1$ and $\theta_2$. These initial values are chosen by the user so that the corresponding configuration already places the EE outside the tetrahedral structure. During the optimization, the algorithm is free to modify the lengths, thicknesses, and angles, but, being a local search method, it tends to converge to a solution that remains in the proximity of the initial guess.
\\
By initializing the design variables with a configuration that satisfies the clinical requirements on line of sight, rapid EE exchange, and sterilizability, we bias the optimization towards configurations that preserve an external EE while refining the geometric parameters to meet the mechanical performance targets.


\paragraph{Compliance formulation}
We formulate a planar compliance model with node $N1$ fixed and compliance evaluated at node $N4$. First we define the $2\times2$ translational compliance matrix $\mathbf{S}$
\begin{equation}
\mathbf{S}=
\begin{bmatrix}
S_{11} & S_{12} \\
S_{21} & S_{22}
\end{bmatrix},
\label{eq:matrix_S}
\end{equation}
where $S_{11}$ is the horizontal displacement $u_x$ under a unit horizontal force $F_x$, $S_{12}$ the horizontal displacement under a unit vertical force $F_y$, $S_{21}$ the vertical displacement $u_y$ under a unit horizontal force, and $S_{22}$ the vertical displacement under a unit vertical force $F_y$.

Each panel $p_i$ is approximated as an equivalent slender beam with rectangular cross-section, with height $b$, thickness $t_i$, and length $L_i$. The cross-sectional area $\mathbf{A}_i$ and the second moment $\mathbf{I}_i$ of the area are
\begin{equation}
\mathbf{A}_i=bt_i, \qquad
\mathbf{I}_i=\frac{b\,t_i^3}{12},
\label{eq:SectionInertia}
\end{equation}
and the axial $\mathbf{K}_{N,i}$ and bending stiffnesses $\mathbf{K}_{V,i}$ are
\begin{equation}
\mathbf{K}_{N,i}=\frac{E A_i}{L_i}, \qquad
\mathbf{K}_{V,i}=\frac{12 E I_i}{L_i^3},
\label{eq:KNKV}
\end{equation}
with $E$ the Young’s modulus. The in-plane stiffness of the three-panels subsystem is then obtained using the stiffness matrix $\mathbf{K}$:
\begin{equation}
\mathbf{K}=
\begin{bmatrix}
K_{xx} & K_{xy}\\
K_{yx} & K_{yy}
\end{bmatrix},
\label{eq:matrix_K}
\end{equation}
with components
\begin{align}
K_{xx} &= \sum_{i=1}^{3}\!\left( K_{N,i}\cos^2\theta_i + K_{V,i}\sin^2\theta_i \right), \nonumber\\
K_{yy} &= \sum_{i=1}^{3}\!\left( K_{N,i}\sin^2\theta_i + K_{V,i}\cos^2\theta_i \right), \label{eq:Kentries}\\
K_{xy} &= K_{yx}=\sum_{i=1}^{3}\!\left( K_{N,i}-K_{V,i} \right)\cos\theta_i\sin\theta_i, \nonumber
\end{align}
where $\theta_3$ is obtained by the sum $\theta_1+\theta_2$. 

The compliance is $\mathbf{S}=\mathbf{K}^{-1}$, that is corresponding to,
\begin{equation}
\mathbf{S}
=\frac{1}{\det(\mathbf{K})}
\begin{bmatrix}
K_{yy} & -K_{xy}\\
-K_{yx} & K_{xx}
\end{bmatrix},
\label{eq:matrix_S_from_K}
\end{equation}
so that
\begin{align} S_{11} &= \frac{K_{yy}}{\det(\mathbf{K})}, \\ S_{12} &= -\frac{K_{xy}}{\det(\mathbf{K})}, \\ S_{21} &= -\frac{K_{yx}}{\det(\mathbf{K})}, \\ S_{22} &= \frac{K_{xx}}{\det(\mathbf{K})}. \end{align}

For an isotropic planar response, the compliance matrix must be proportional to the identity, $S=sI$, implying $S_{11}=S_{22}$ and vanishing coupling terms $S_{12}=S_{21}=0$. We therefore introduce the following anisotropy index, $idx$, which quantifies the deviation from these conditions by combining the diagonal mismatch and off-diagonal terms, normalized by a representative compliance magnitude.

\begin{equation}
\mathrm{idx}=
\frac{\sqrt{(S_{11}-S_{22})^2+S_{12}^2+S_{21}^2}}
{\mathrm{mean}(S_{11},S_{12},S_{21},S_{22})},
\label{eq:anisotropy_index}
\end{equation}
which is minimized by iteratively updating $\{L_1,L_2,L_3\}$, $\{t_1, t_2,t_3\}$, and $\{\theta_1,\theta_2\}$ toward $\mathrm{idx}\to 0$.

\paragraph{Scale invariance and parameter reporting}
Because the response depends primarily on ratios of the geometric variables, the problem is scale-invariant. Therefore, lengths are reported normalized by $\min(L_i)=L_{ref}$ and thicknesses by $\min(t_i)=t_{ref}$, allowing an absolute choice for length and thickness while determining the remaining parameters parametrically. These two values will be used for the parametric model of \ref{stiffness}. Angles $\theta_1$ and $\theta_2$ are explicitly reported.

\paragraph{Finite-element verification}
From this section onward, the joint is evaluated using the full 3D FEM model rather than the equivalent-beam approximation.

The analytical model is a reduced-order formulation introduced for tractability to derive a closed-form isotropy criterion and enable rapid design exploration. By contrast, the FEM model is built from the full 3D CAD geometry of the complete joint, and therefore does not rely on the same geometric simplifications. The analysis is carried out in the linear elastic, small-deflection regime and uses the same loading direction representative of operation.

The analytical predictions were verified in \textsc{Ansys} Workbench 2023 R1 (Ansys, Inc) by finite element analysis of the three-panels subsystem. With $N1$ fixed, a force of magnitude $\mathbf{F}$ was applied at $N4$ and its direction rotated by an angle $\theta_f$ about $y-axis$ (Fig.~\ref{fig:scheme_tretravi}) from $0^\circ$ to $360^\circ$ in increments of $10^\circ$. We computed the directional stiffness as
\begin{equation}
k(\theta_f)=\frac{F}{x(\theta_f)},
\label{eq_stiffness}
\end{equation}
where $x(\theta_f)$ is the displacement magnitude at $N4$. Since $k$ scales linearly with $F$, the absolute value of $F$ is immaterial for this study.

\paragraph{Ellipse fitting and principal-axis ratio}
The $k(\theta_f)$ polar plot is approximately elliptical. We estimate stiffness as a function of orientation over $0$–$360^\circ$ by fitting an ellipse to the sampled data using the direct least squares method \cite{fitzgibbon1996direct}. The method: (i) enforces the ellipse condition $B^2-4AC<0$ a posteriori to ensure that the fitted conic is indeed an ellipse; (ii) extracts geometric parameters (center and semi-axes); and (iii) overlays the fitted ellipse on the raw data for visual inspection. The proximity to isotropy is quantified by the Principal Axis Ratio (PAR)
\begin{equation}
\mathrm{PAR}=\frac{A}{B},
\label{eq:par}
\end{equation}
with $A$ and $B$ the major and minor semiaxes of the ellipse, respectively. The perfect isotropy corresponds to $\mathrm{PAR}=1$, that is a circle.

\subsection{Design of Constraining Panels for RCM Rotation Suppression}
\label{stiffness}

The \enquote{red} panels in Fig.~\ref{fig:walls_tetraII} complete the tetrahedral assembly and suppress rotation about the RCM, producing the optimal joint configuration.

A parametric model was implemented in \textsc{Ansys}, with explicit input and output parameters defined as follows.

\paragraph{Input parameters}
\begin{itemize}
\item \textbf{Minimum panel length ($L_{ref}$):} All panel lengths are parameterized from a single reference (see Subsection~\ref{isotropic}). $L_{ref}$ is treated as independent, and the remaining lengths follow from this parameterization.
\item \textbf{Joint height ($H$):} Overall joint height (Fig.~\ref{fig:Joint_3D}.b).
\item \textbf{Minimum panel thickness ($t_{ref}$):} By analogy with length, the panel thicknesses $t_1$, $t_2$, and $t_3$ are parameterized by $t_{ref}$.
\item \textbf{Triangular-structure thickness ($t_{\triangle}$):} thickness of the triangular substructure that constrains the RCM during the EE movement (Fig.~\ref{fig:Joint_3D}.a).
\item \textbf{Angle $\boldsymbol{\alpha}$:} Angle between the red and blue panels (Fig.~\ref{fig:Joint_3D}.a). To reduce dimensionality, all such angles were set equal in the simulations.
\end{itemize}

\paragraph{Output parameters}
\begin{itemize}
\item \textbf{EE displacement:} Magnitude of EE displacement measured at N4 with $\theta_f\!=\!0^\circ,\,120^\circ,\,240^\circ$ ($r_0$, $r_{120}$, $r_{240}$).
\item \textbf{Parasitic RCM error:} Parasitic displacement in the RCM under EE motion, measured in the same directions ($err_0$, $err_{120}$, $err_{240}$).
\end{itemize}

Three load cases were analyzed applying a force at $N4$ with orientation $\theta_f\in\{0^\circ,120^\circ,240^\circ\}$. Node $N1$ was fixed. The following performance metrics were then evaluated during the feasibility analysis:
\begin{equation}
\mathrm{IsoErr}=\frac{\max(r_0, r_{120}, r_{240})-\min(r_0, r_{120}, r_{240})}{\mathrm{mean}(r_0, r_{120}, r_{240})},
\label{eq_isoerror}
\end{equation}
\begin{equation}
\mathrm{ParErr}=\frac{r_0+r_{120}+r_{240}}{\,err_0+err_{120}+err_{240}\,}.
\label{eq_parerr}
\end{equation}
Here, $\mathrm{IsoErr}$ is dimensionless and attains its minimum ($0$) when the three radii coincide, i.e., when workspace displacement (hence stiffness response) is isotropic. $\mathrm{ParErr}$ is the ratio of cumulative EE displacement to cumulative parasitic RCM error; larger values indicate better suppression of parasitic motion for a given mobility.

In a planar field, anisotropy is associated with ellipse-like directional behavior, which is fully described by two principal axes. A third, non-collinear sample is the theoretical minimum needed to reveal orientation-dependent variation. The uniform spacing at $120^\circ$ maximizes angular coverage with a minimal rotation invariant test set.

\begin{figure*}
    \centering
    \includegraphics[width=1\linewidth]{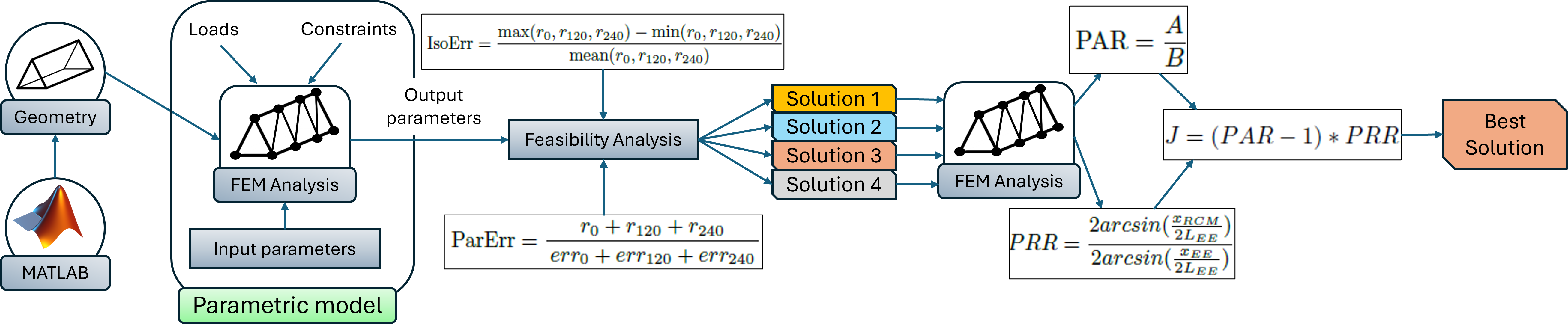}
    \caption{Workflow of the study: MATLAB identifies candidate geometries that satisfy the isotropic-stiffness condition for the mobility panels. A parametric FEM model is then evaluated and the metrics $IsoErr$ and $ParErr$ are computed from the output displacements. A feasibility analysis selects admissible designs and returns Solutions 1–4. These candidates are re-simulated with FEM, and the best configuration is chosen using the Principal-Axis Ratio $PAR$, the Parasitic-to-useful Rotation Ratio $PRR$, and the composite index $J$.}
    \label{fig:Ansys_project}
\end{figure*}

A feasibility study, shown in Fig. \ref{fig:Ansys_project}, was conducted by sweeping the input parameters over prescribed limits (Table~\ref{tab:parametri_range}) and randomly perturbing them up to 500 times. The chosen material was DuraForm Polyamide~12 (PA12), with Young’s modulus $E\approx 1{,}400~\mathrm{MPa}$ and elongation at break $\approx14\%$, suitable for compliant mechanisms. The analysis was carried out modeling PA12 as an isotropic linear-elastic material. The joint was discretized with 3D solid tetrahedral elements and a maximum element size of 2 mm, selected to adequately resolve the compliant panels and capture bending-dominated deformation. Because the problem is bi-objective (minimize IsoErr and maximize ParErr), the exploration does not necessarily return a unique optimum. Instead, it can produce a set of candidate designs in which improving one objective would deteriorate the other. For example, some configurations achieve very low IsoErr at the expense of a less favorable ParErr, while others reduce ParErr with a higher IsoErr.

\begin{table}[h!]
\centering
\caption{Parameter ranges}
\label{tab:parametri_range}
\begin{tabular}{lll}
\hline
\textbf{Parameter} & \textbf{Type} & \textbf{Interval} \\
\hline
$L_{ref}$         & Length        & $20\,\mathrm{mm} \le L_{ref} \le 100\,\mathrm{mm}$ \\
$t_{ref}$         & Thickness     & $1\,\mathrm{mm} \le t_{ref} \le 6\,\mathrm{mm}$ \\
$t_{\triangle}$    & Thickness     & $1\,\mathrm{mm} \le t_{\triangle} \le 6\,\mathrm{mm}$ \\
$H$                & Height        & $20\,\mathrm{mm} \le H \le 100\,\mathrm{mm}$ \\
$\alpha$           & Angle         & $10^{\circ} \le \alpha \le 70^{\circ}$ \\
\hline
\end{tabular}
\end{table}

\newcommand{\panelheight}{8cm}

\begin{figure*}[t]
    \centering
    \begin{subfigure}[t]{0.30\textwidth}
        \centering
        \includegraphics[height=\panelheight,keepaspectratio]{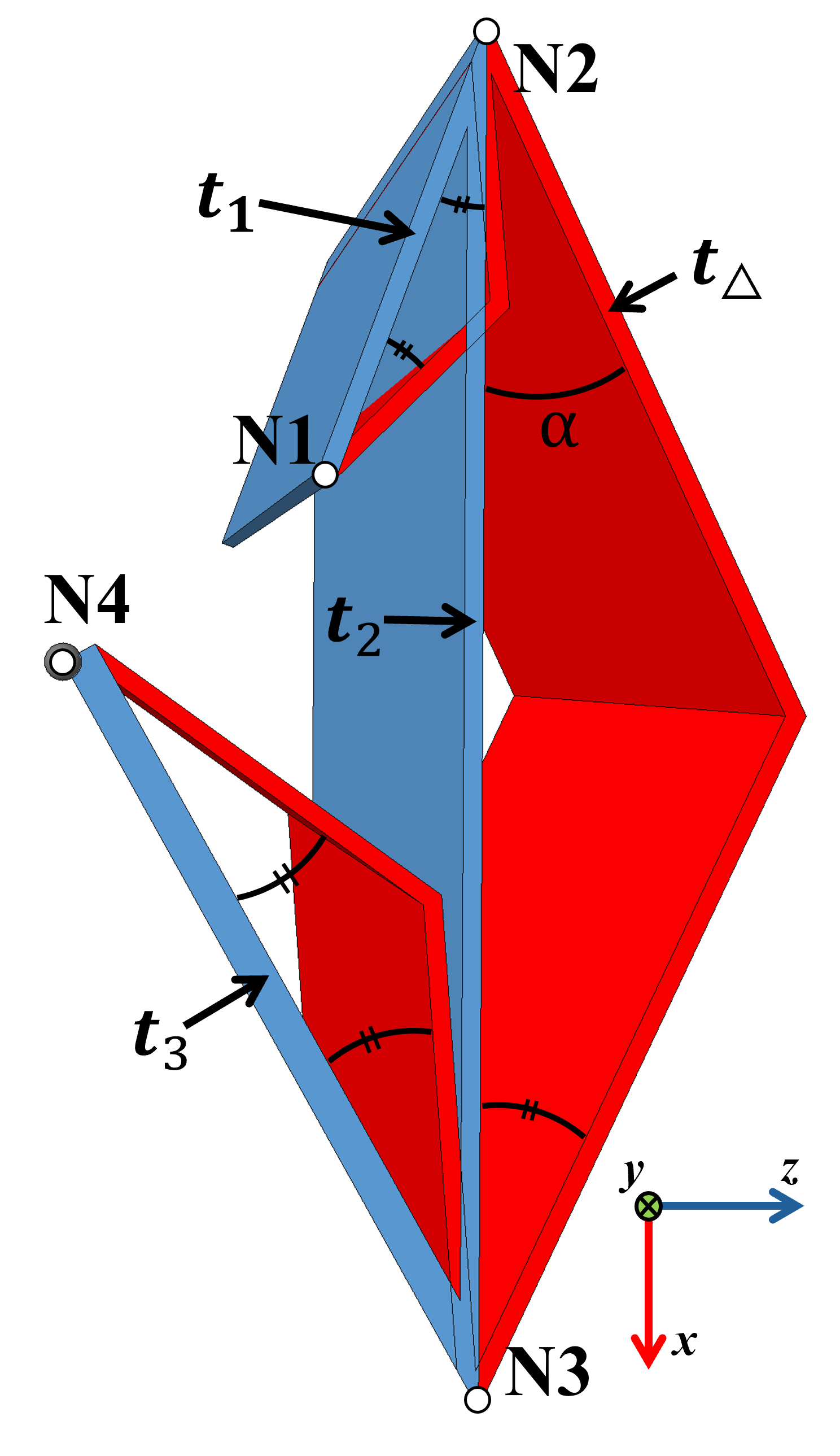}
        \caption{}
        \label{fig:proto}
    \end{subfigure}\hfill
    \begin{subfigure}[t]{0.30\textwidth}
        \centering
        \includegraphics[height=\panelheight,keepaspectratio]{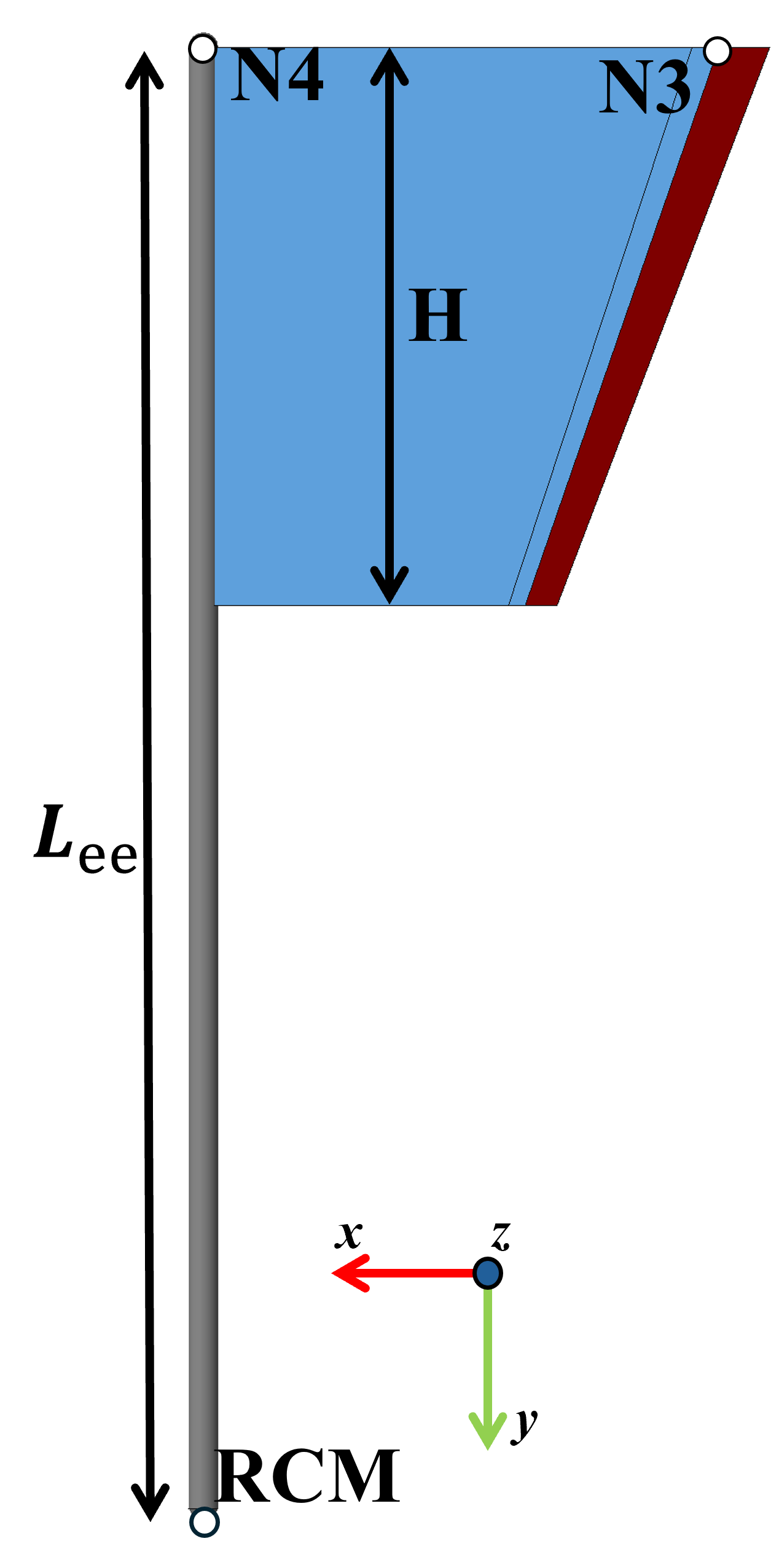}
        \caption{}
        \label{fig:stiffness}
    \end{subfigure}\hfill
    \begin{subfigure}[t]{0.30\textwidth}
        \centering
        \includegraphics[height=\panelheight,keepaspectratio]{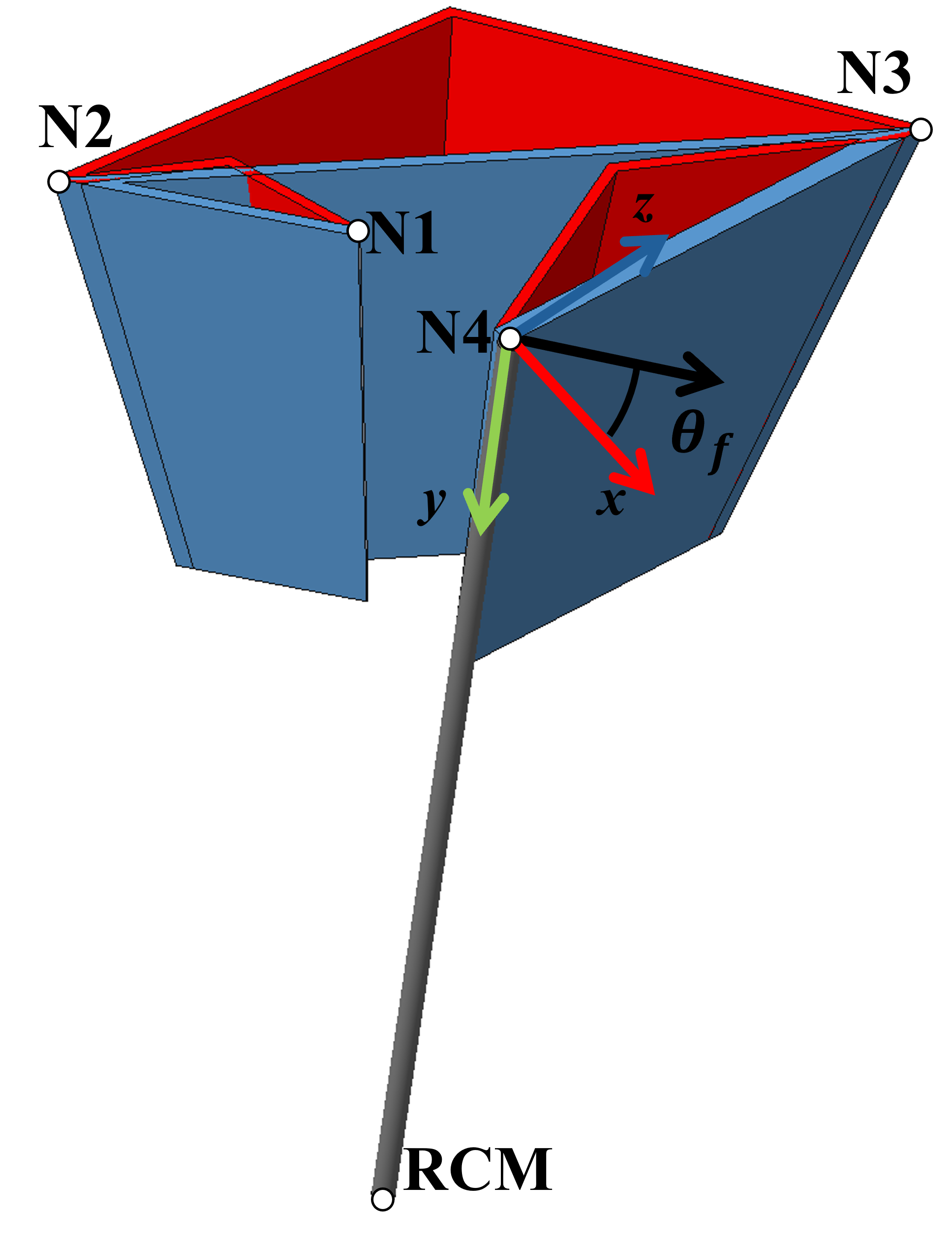}
        \caption{}
        \label{fig:third}
    \end{subfigure}
    \caption{Representation of the joint from frontal view in \textbf{(a)}, lateral view in \textbf{(b)} and isometric view in \textbf{(c)}. In blue, the structure is responsible for the movement of the end effector, and in red, the constraining structure locks the RCM during the movement. Highlighted are the nodes $N1, N2, N3$ and $N4$ with the angle $\alpha$ and all the thickness $t1, t2, t3$ and $t_\triangle$.}
    \label{fig:Joint_3D}
\end{figure*}

\subsection{Optimization of the Joint Configuration} \label{Optimized}
The best configurations obtained in the previous subsection were then tested separately by analyzing the displacement obtained by rotating a force at N4 while maintaining the fixed point N1. The force was rotated around $\theta_f$ at $360^\circ$ with an interval between each observation of $10^\circ$.

This phase was performed to obtain the displacement of node N4 and the displacement of the RCM point, which corresponds to the parasitic error.
The displacement values obtained at the node N4 were then converted using Eq. (\ref{eq_stiffness}) and tsubsequently used for ellipse fitting as explained in Subsection \ref{isotropic}. 

For finding the best solution, we decided to use two objective functions, as shown in Fig. \ref{fig:Ansys_project}: 
\begin{itemize}
    \item \textbf{PAR:} Again we decided to use the Eq. (\ref{eq:par}) for analyzing the isotropy condition. 
    \item \textbf{Parasitic-to-useful Rotation Ratio (PRR):} To quantify how much of the commanded motion is “wasted” in an unwanted shift of the RCM, we introduce the Parasitic-to-useful Rotation Ratio
    \begin{equation}
        PRR=\frac{2arcsin(\frac{x_{RCM}}{2L_{EE}})}{2arcsin(\frac{x_{EE}}{2L_{EE}})},
        \label{eq_PRR}
    \end{equation}
where $L_{EE}$ is the length of the EE, $x_{RCM}$ is the linear drift of the RCM and $x_{EE}$ is the linear displacement of the EE at a distance $L_{EE}$ from the RCM.
\end{itemize}

A value of $\mathrm{PAR}\!\approx\!1$ indicates that the ellipse approaches a circle, i.e. the response is nearly isotropic and the $\mathrm{PRR}\!\approx\!0$ rotation ratio approaches $0\%$, indicating negligible RCM drift.
Furthermore, to obtain an absolute safety indicator, we also evaluated parasitic displacement by imposing a fixed rotation of $4.5^\circ$ at the point N4, which is necessary to move the EE inside the patient’s brain, and measuring the resulting RCM drift. Configurations with $|x_{RCM}|<1,\mathrm{mm}$ were classified as \textit{safe}, following the clinical requirements. In contrast, those with $|x_{RCM}|\geq 1,\mathrm{mm}$ were deemed \textit{unsafe}.

Since the analysis is bi-objective, the feasible set may contain multiple solutions. To select a single optimal configuration, we combine the two performance indices into a scalar objective using
\begin{equation}
    J=(PAR-1)*PRR,
    \label{eq_J}
\end{equation}
and seek the solution for which $J$ is as close to zero as possible.

\subsection{Maximum Workspace}
\label{workspace}

For intraoperative applicability, the attainable workspace is critical. The minimum requirement to steer the EE within the brain and to superimpose its trajectory on the surgeon-defined path is at least $\pm 15^\circ$. In earlier designs, the workspace was estimated by advancing the EE until self-contact produced a sharp stiffness increase. In the present geometry, panel contact may not occur, or may occur only for large EE excursions, therefore, the maximum admissible displacement is determined by fatigue-based joint assessment.

A Wöhler (S–N) curve for PA12 relates the stress amplitude $S$ to the number of cycles to failure $N$ under constant-amplitude loading. Following Salazar \emph{et~al.}~\cite{salazar2022mechanical}, the S–N relation for SLS-manufactured PA12 with $0^\circ$ build orientation is:
\begin{equation}
S = 111.1\, N^{-0.11}.
\label{eq_SNcurve}
\end{equation}
The displacement of the EE increases until the safety factor $\mathrm{SF}$ reaches a prescribed threshold. If $\mathrm{SF}>1$, the joint is expected to sustain cyclic operation without failure. The admissibility limit is taken as $\mathrm{SF}=1$.

The maximum EE rotation about the RCM corresponding to $\mathrm{SF}=1$ is recorded while sweeping the actuation direction $\theta_f$ from $0^\circ$ to $360^\circ$ in $10^\circ$ increments. Let $\mathbf{d}_{\mathrm{WS}}(\theta_f)$
 denote the displacement at the threshold obtained at a specific value of $\theta_f$, and let $L_{\mathrm{EE}}$ be the EE shaft length, the corresponding angular excursion is
\begin{equation}
\bm{\beta}_{\mathrm{WS}}(\theta_f) = 2\,\arctan\!\left(\frac{\mathbf{d}_{\mathrm{WS}}(\theta_f)}{2\,L_{\mathrm{EE}}}\right).
\end{equation}
The resulting $\bm{\beta}_{\mathrm{WS}}(\theta_f)$ characterizes the maximum fatigue-safe workspace of the joint.

To support the linear-elastic FEM assumption adopted in this work, we evaluated the von Mises equivalent stress under the nominal workspace motion while sweeping the actuation direction, following the same procedure used for the fatigue analysis. A necessary condition to remain in the elastic regime is that the peak von Mises stress stays below the tensile yield stress, reported for dry PA12 samples at $23^\circ$C at $45.5~MPa$ \cite{amstutz2021temperature}.


\subsection{Validation of the Stiffness Behavior} \label{validation}
The purpose of the validation was to compare the stiffness values obtained on the redesigned off-axis RCM joint obtained in the simulation with an experiment simulating the same scenario.
A step-by-step protocol is detailed below.

A monolithic joint prototype was produced in SLS using PA12 as the primary material. PA12 was selected for its bio-compatible characteristics, Young’s modulus of 1.4 GPa, and elongation-at-break of 14\%. 

A 6 DOF Aurora sensor (Northern Digital Inc.)
, was used to evaluate the movement of point N4. 
The Aurora field generator sequentially energizes the 
sensor 
solving in real time its absolute position and orientation. 
Accurate tracking is maintained without line of sight, even amid metallic instruments and other clutter commonly found in surgical settings \cite{yaniv2009electromagnetic}.

The Aurora tracking system does not allow a direct experimental validation of the parasitic error: the manufacturer specifies a positional accuracy of $1.40$ mm and an orientation accuracy of $0.35^\circ$ ($95\%$ CI), whereas the predicted RCM shift is expected to remain below 1 mm. Therefore, in this work the parasitic error is estimated using only the FEM simulations.

To emulate the typical manipulation forces imparted by a neurosurgeon during endoscope steering, a weight of $204\pm1~g$, generating a force of $\approx2N$,  was suspended from point N4. Because stiffness is evaluated as $k=F/x$ and the joint response is linear within the tested range, the load magnitude does not affect the stiffness estimate. A 2 N load was chosen as a compromise to produce a measurable displacement above sensor noise and setup parasitic effects in the setup, while avoiding excessive stresses or risk of damage of the PA12 flexures.

The joint was clamped at its base (N1) on a rotary stage and incrementally rotated about the y-axis in $30^\circ$ steps over a full $360^\circ$. Ten consecutive displacement readings were captured per orientation, resulting in 120 samples.

Cartesian displacements were transformed into polar stiffness, $\mathbf{k}$ using Eq. (\ref{eq_stiffness}).

We defined also the percentage error between the value obtained during the simulation, explained in Section \ref{Optimized} and during the experiment using the following equation: 
\begin{equation}
    \% Error=\frac{|Exp-Sim|}{Exp},
\end{equation}
where $Exp$ are the stiffness values obtained during the experimental validation and $Sim$ are the stiffness values obtained during the simulation.

For each of the 12 tested orientations, the 10 repeated measurements were summarized as mean $\pm$ standard deviation. The agreement between the experimental mean values and the corresponding simulated stiffness values was then quantified over the 12 tested orientations through the mean absolute error (MAE), root mean square error (RMSE), mean absolute percentage error (MAPE), and mean bias.

\begin{figure}
    \centering
    \includegraphics[width=1\linewidth]{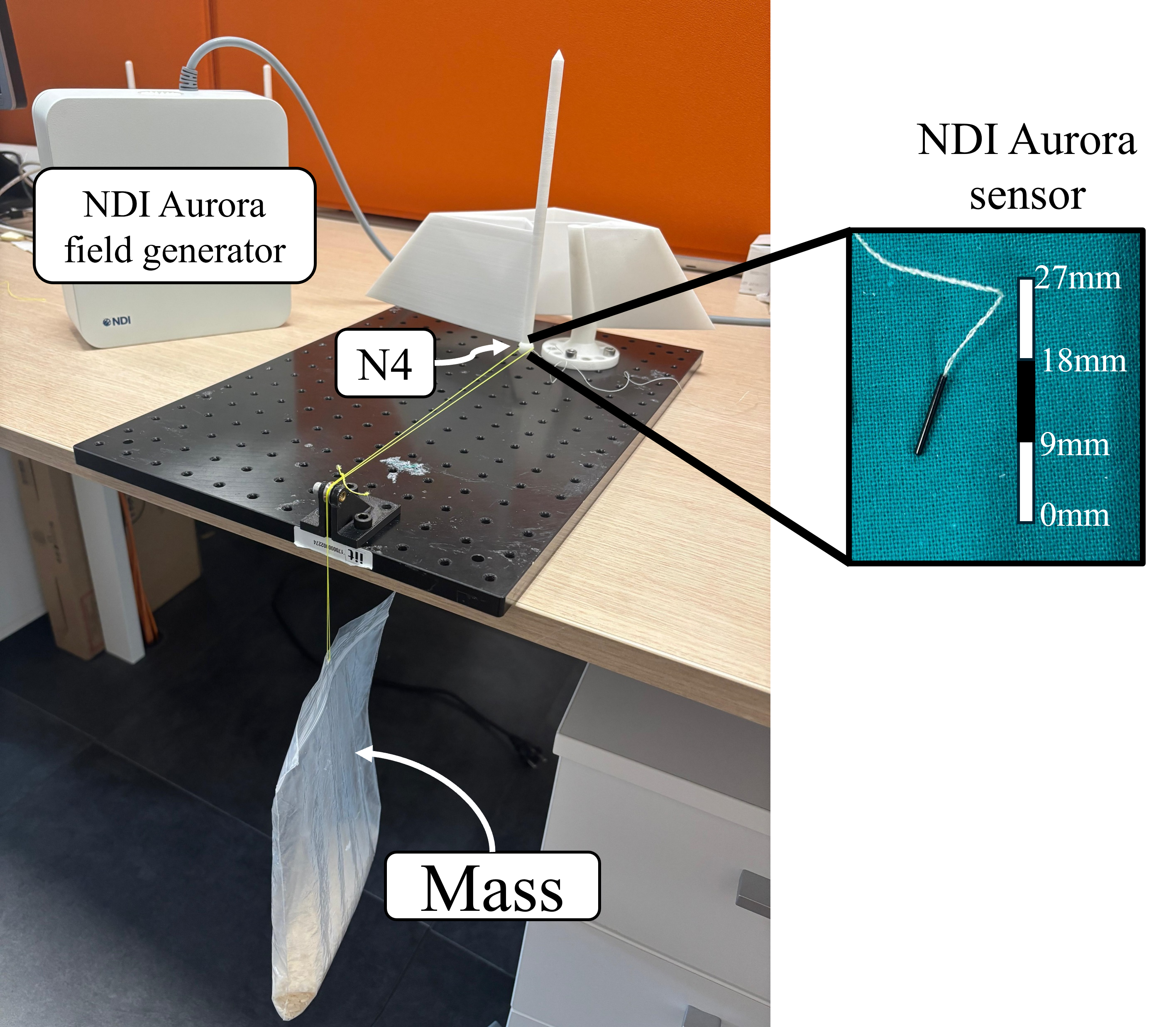}
    \caption{Experimental setup used to validate the joint in real-life conditions. A mass producing a force of $2N$ is attached at point N4, while an Aurora microsensor, operating within the Aurora electromagnetic field, is used to measure the movement of point N4.}
    \label{fig:placeholder}
\end{figure}

\begin{figure*}[t]
    \centering
    \begin{subfigure}[t]{0.3\textwidth}
        \centering
        \includegraphics[width=\linewidth]{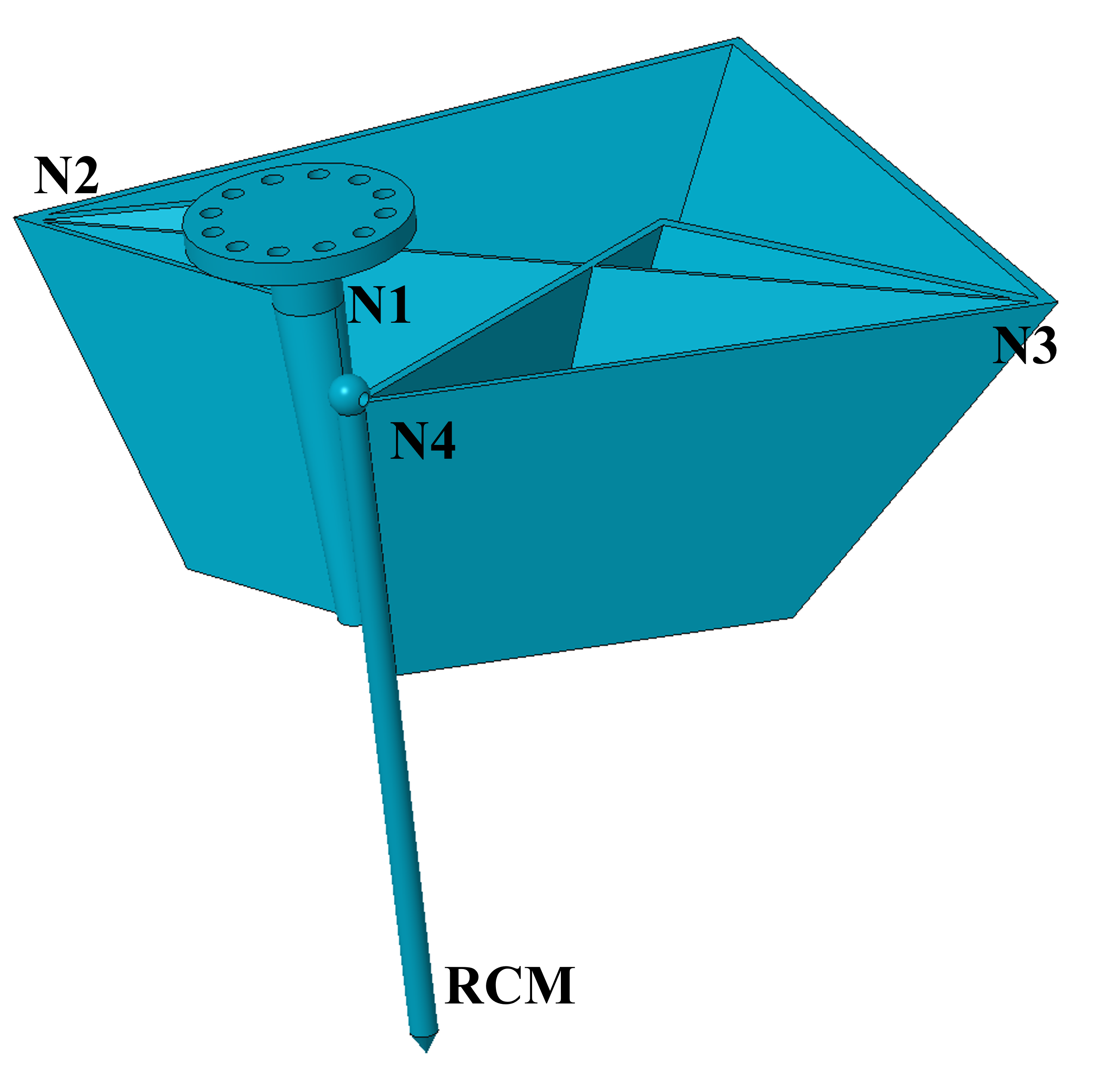}
        \caption{}
        \label{fig:proto}
    \end{subfigure}\hfill
    \begin{subfigure}[t]{0.35\textwidth}
        \centering
        \includegraphics[width=0.9\linewidth]{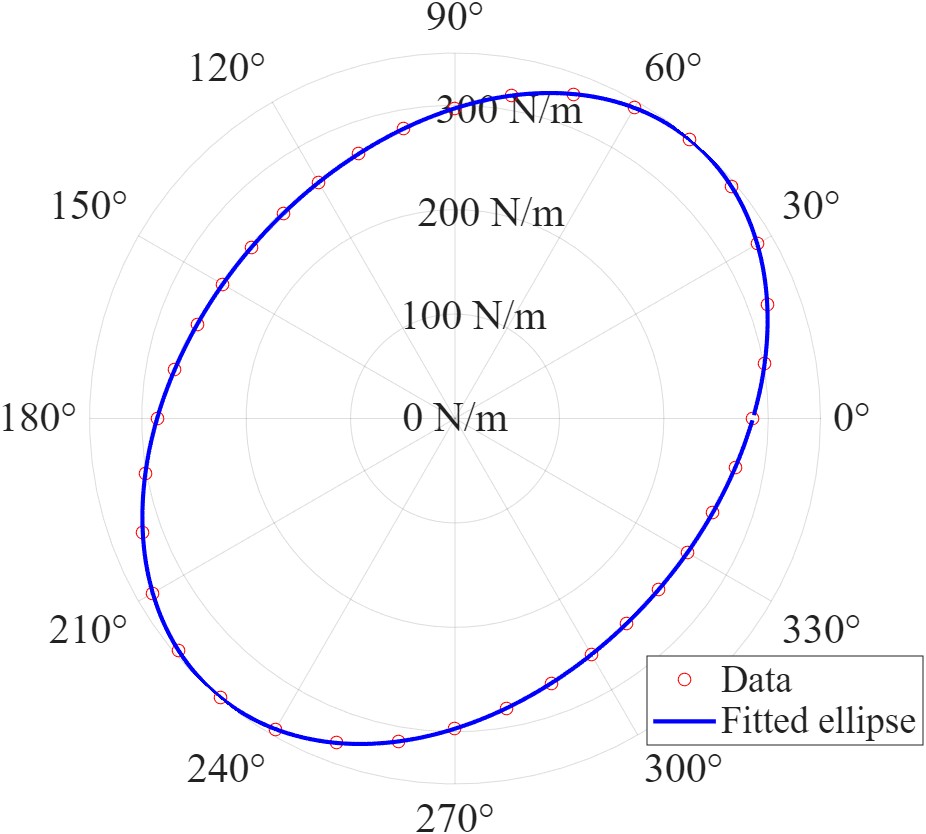}
        \caption{}
        \label{fig:stiffness}
    \end{subfigure}\hfill
    \begin{subfigure}[t]{0.3\textwidth}
        \centering
        \includegraphics[width=\linewidth]{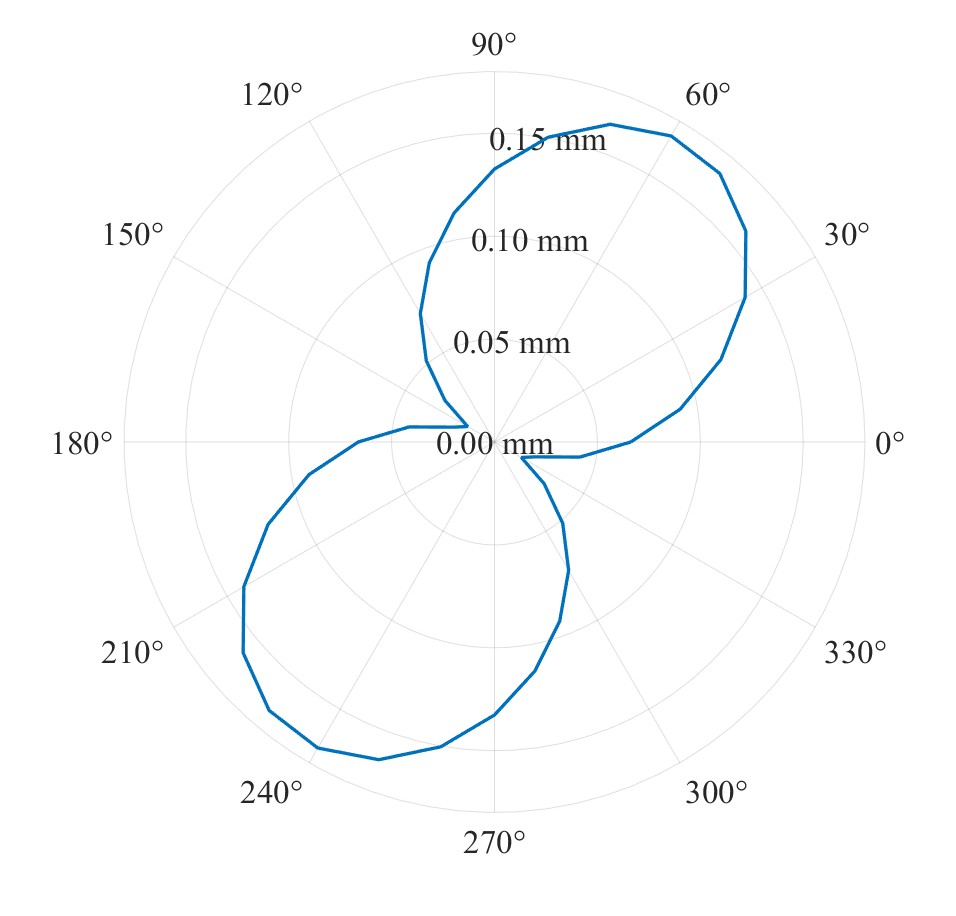} 
        \caption{}
        \label{fig:third}
    \end{subfigure}
    \caption{\textbf{(a)} 3D model of the final compliant joint. \textbf{(b)} Polar graph showing} the stiffness obtained around the $\theta_f$ for the final compliant joint after the feasibility analysis. The red circle corresponds to simulation data, while the blue line is the ellipse fitted in MATLAB. \textbf{(c)} Polar graph of the parasitic error obtained while moving the EE of $4.5^\circ$ around $\theta_f$.
    \label{fig:sn_side_by_side}
\end{figure*}

\section{Results}
\subsection{Results Analytical Identification of the Isotropic Stiffness Criterion}
To guarantee direction-independent rigidity, the panel geometry must satisfy the scale-invariant ratios derived in Subsection~\ref{isotropic}. MATLAB analysis identified $L_1$ as the minimum length, so we set the reference $L_{ref}=L_1$. Similarly, $t_2$ was obtained as the minimum thickness and was taken as the reference $t_{ref}=t_2$. The other dimensions were then obtained as parameters of $L_{ref}$ and $t_{ref}$. 

\begin{table}[t]
\centering
\caption{Parameters of the isotropic stiffness solution.}
\label{tab:isotropic_params}
\begin{tabular}{lll}
\toprule
\textbf{Parameter} & \textbf{Definition} & \textbf{Value / Ratio} \\
\midrule
\(L_1\) & reference length & \(L_{ref}\) \\
\(L_2/L_1\) & --- & 3.12 \\
\(L_3/L_1\) & --- & 2.00 \\
\(t_2\) & reference thickness & \(t_{ref}\) \\
\(t_1/t_2\) & --- & 1.03 \\
\(t_3/t_2\) & --- & 2.01 \\
\(\theta_1\) & panel-1 orientation & \(159.8^{\circ}\) \\
\(\theta_2\) & panel-2 orientation & \(150.6^{\circ}\) \\
\bottomrule
\end{tabular}
\end{table}

The length $L_{ref}$ and the thickness $t_{ref}$ can be freely chosen, and the other lengths and thicknesses will be modified following the relations illustrated in Table \ref{tab:isotropic_params}, maintaining the same anisotropy index.

The stiffness verification was performed on ANSYS Workbench 2023 R1 simulating three cases with $L_{ref}$=20 mm, $t_{ref}$=1mm, the second case $L_{ref}$=40mm, $t_{ref}$=1mm and then $L_{ref}$=40mm, $t_{ref}$=2mm.
The analyzes were performed by simulating a force at node N4 rotating it around the angle $\theta_f$ at $360^\circ$. The angle was varied between each observation of $10^\circ$ step. 
Knowing that the graph of the stiffness behavior is elliptical, we used MATLAB software to obtain the elliptical graph of the stiffness. Then we used Eq. (\ref{eq:par}) to analyze how close we are to the isotropic condition. 
The results are shown in Table \ref{tab:ansys_iso}.

\begin{table}[t]
\centering
\caption{Verification of isotropic stiffness via ANSYS for three geometric scalings.}
\label{tab:ansys_iso}
\sisetup{table-number-alignment=center,round-mode=places,round-precision=2}
\begin{tabular}{
l
S[table-format=2.0] 
S[table-format=1.0] 
S[table-format=5.0] 
S[table-format=5.0] 
S[table-format=5.0] 
S[table-format=1.2] 
S[table-format=1.3] 
}
\toprule
\textbf{Case} & {$L_{ref}$ [mm]} & {$t_{ref}$ [mm]} & PAR \\
\midrule
1 & 20 & 1 & {1.12} \\
2 & 40 & 1 & {1.06} \\
3 & 40 & 2 & {1.10} \\
\bottomrule
\end{tabular}

\vspace{2mm}
\end{table}

\subsection{Results Design of Constraining Panels for RCM Rotation Suppression}
In Subsection \ref{stiffness} we add the torsion stiffeners, defined also as \enquote{red} panels, to fix the tip of the EE to rotate around the RCM. A feasibility analysis was defined simulating a force at the node N4 at three different orientation of $\theta_f$ at $0^\circ, 120^\circ$ and $240^\circ$ using as input parameters $L_{ref}$, $H$, $t_{ref}$, $t_\triangle$ and $\alpha$. The objective is to reduce the value of $IsoErr$, Eq. (\ref{eq_isoerror}), and increase the value of $ParErr$, Eq. (\ref{eq_parerr}). The feasibility gives us four configurations, shown in Table \ref{tab:configurazioni}, among the 500 analyzed.

\begin{table}[h!]
\centering
\caption{configuration obtained from the feasibility analysis}
\begin{tabular}{lcccc}
\hline
 & Conf. 1 & Conf. 2 & Conf. 3 & Conf. 4 \\
\hline
$t_{ref}$ [mm]      & 1.59   & 1.15   & 1.33    & 1.83 \\
H [mm] & 97.56  & 84.01  & 95.98   & 67.59 \\
$L_{ref}$ [mm]     & 64.10  & 70.84  & 97.86   & 99.76 \\
$t_{\triangle}$ [mm]  & 3.35   & 3.55   & 3.42    & 1.67 \\
$\alpha$ [$^\circ$]        & 34.40  & 46.33  & 42.68   & 26.61 \\
IsoErr     & 0.12   & 0.22   & 0.27    & 0.52 \\
ParErr     & 30.14  & 67.61  & 159.78  & 280.89 \\
\hline
\end{tabular}
\label{tab:configurazioni}
\end{table}

\subsection{Results Optimization of the Joint Configuration}
The optimal solution was found by performing an additional analysis on ANSYS Workbench 2023 R1, simulating a rotating force around the y-axis to find the displacement of the EE in node N4 and the displacement of the EE at the RCM. 
These displacements were then converted into stiffness using Eq. (\ref{eq_stiffness}).


Table~\ref{tab:PAR_PRR_J} reports the results obtained for the four candidate configurations in terms of PAR, Eq.~(\ref{eq:par}), which quantifies the ratio between the major and minor semiaxes of the stiffness ellipse, and PRR, Eq.~(\ref{eq_PRR}), which quantifies the drift of the RCM following a motion of the endoscope at node $N4$. Since the analysis yields four best (non-dominated) configurations, we further combine PAR and PRR into the scalar objective $J$, Eq.~(\ref{eq_J}), and select as optimal the configuration for which $J$ is closest to zero.

\begin{table}[h!]
\centering
\caption{configuration obtained from the feasibility analysis}
\begin{tabular}{lcccc}
\hline
 & Conf. 1 & Conf. 2 & Conf. 3 & Conf. 4 \\
\hline
PAR     & 1.17   & 1.29   & 1.37    & 1.87 \\
PRR & 0.032 & 0.015 & 0.0063 & 0.0041 \\
J    & 0.0055  & 0.0043  & 0.0023  & 0.0036 \\
\hline
\end{tabular}
\label{tab:PAR_PRR_J}
\end{table}

\begin{figure}[t]
  \centering
  \setlength{\tabcolsep}{2pt} 
  \begin{tabular}{@{}cc@{}}
    \subfloat[]{\includegraphics[width=0.45\linewidth]{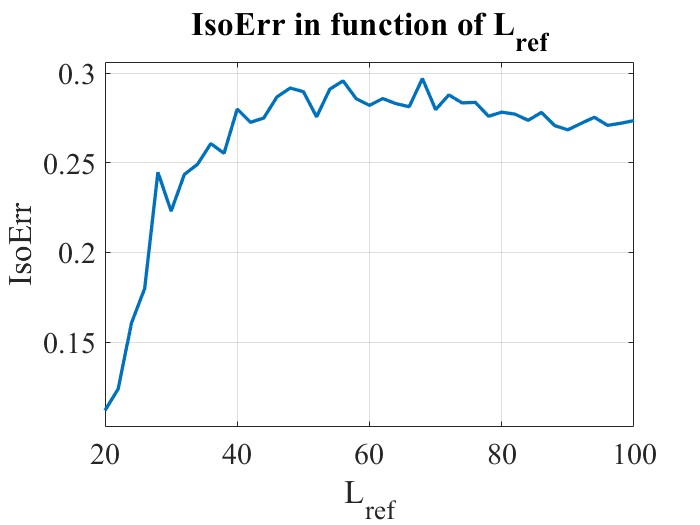}}\ &
    \subfloat[]{\includegraphics[width=0.45\linewidth]{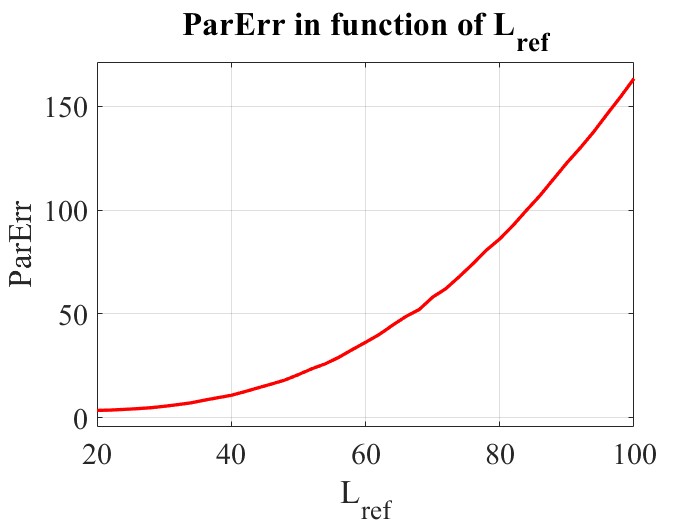}}\\
    \subfloat[]{\includegraphics[width=0.45\linewidth]{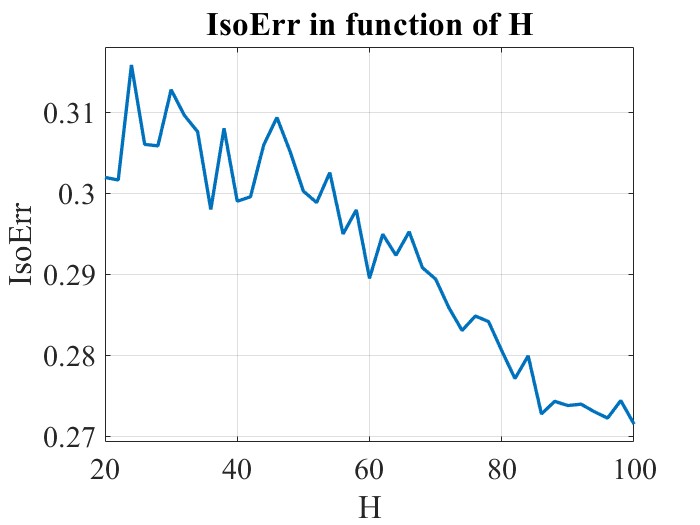}}\ &
    \subfloat[]{\includegraphics[width=0.45\linewidth]{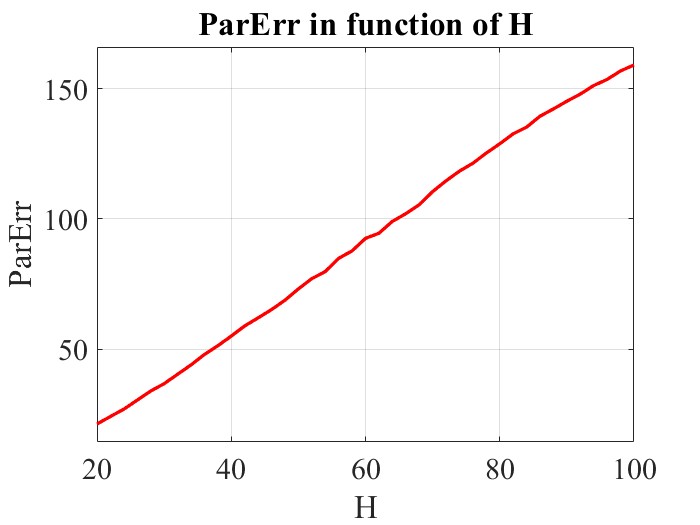}}\\
    \subfloat[]{\includegraphics[width=0.45\linewidth]{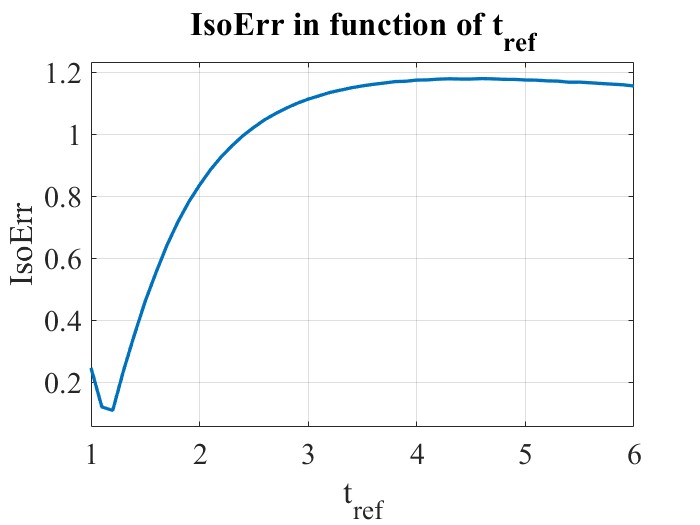}}\ &
    \subfloat[]{\includegraphics[width=0.45\linewidth]{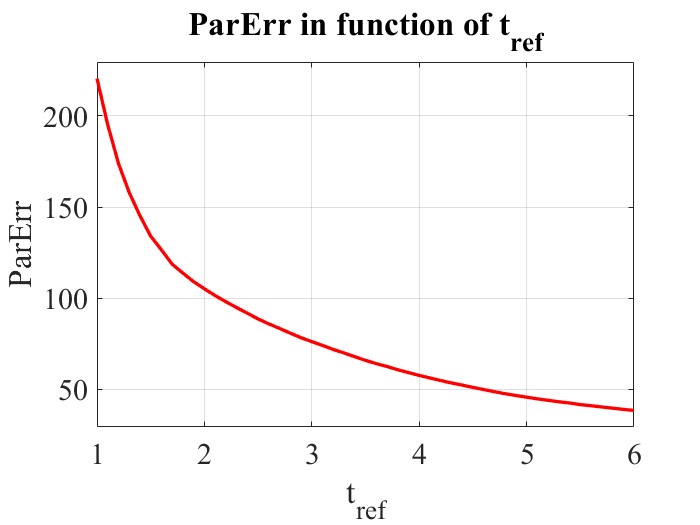}}\\
    \subfloat[]{\includegraphics[width=0.45\linewidth]{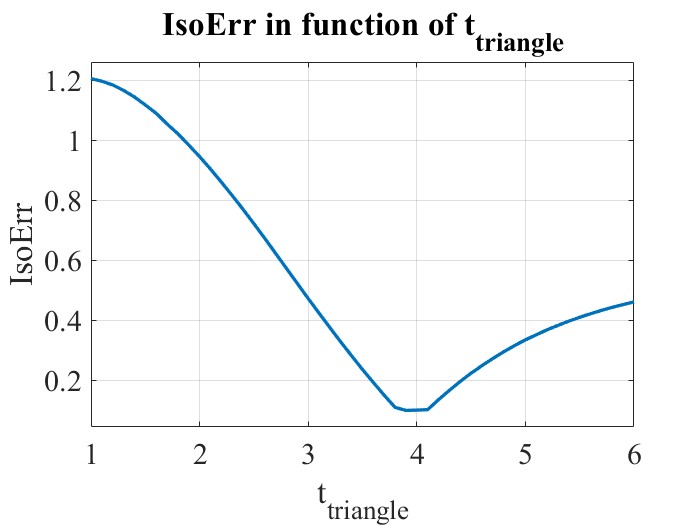}}\ &
    \subfloat[]{\includegraphics[width=0.45\linewidth]{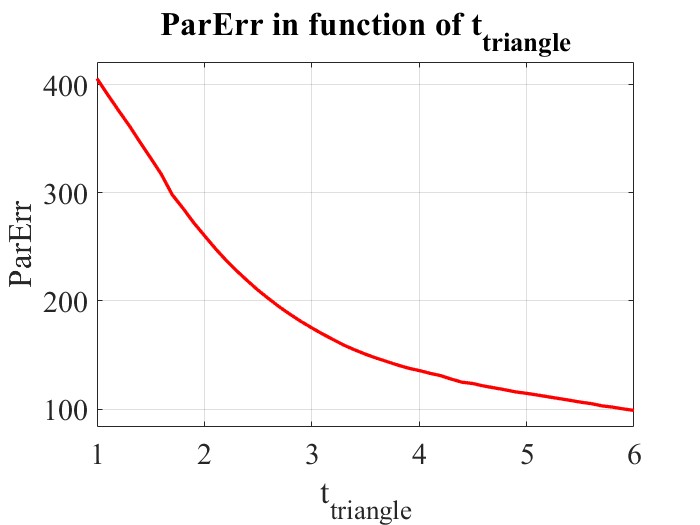}}\\
    \subfloat[]{\includegraphics[width=0.45\linewidth]{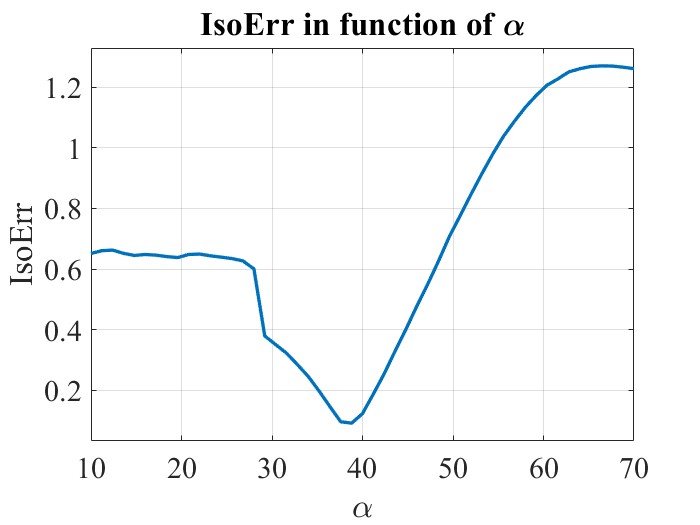}}\ &
    \subfloat[]{\includegraphics[width=0.45\linewidth]{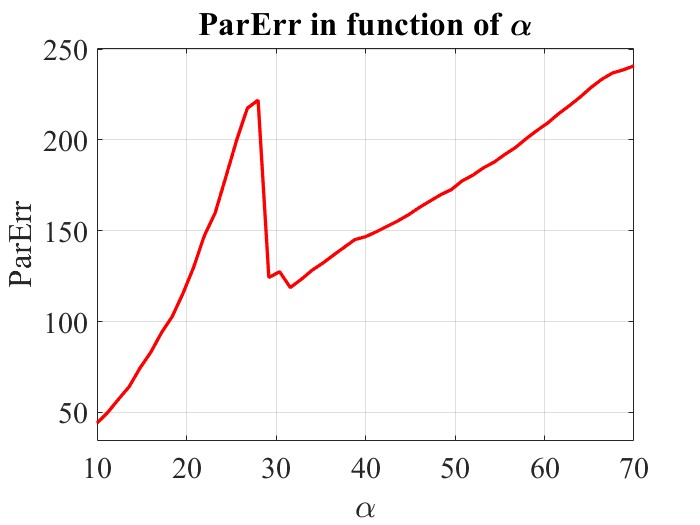}}\\
  \end{tabular}
  \caption{Graphs of two feasibility metrics, IsoErr (left column, in blue) and ParErr (right column, in red) versus the five inputs. (\textbf{a-b}) $L_{ref}$, (\textbf{c-d}) $H$, (\textbf{e-f}) $t_{ref}$, (\textbf{g-h}) $t_\triangle$, and (\textbf{i-j}) $\alpha$. In each graph, the shown variable sweeps its range while the other four are fixed at best configuration obtained after the optimization ($L_{ref}=97.86mm$, $H=95.98mm$, $t_{ref}=1.33$, $t_\triangle=3.42mm$ and $\alpha=42.68^\circ$.}
  \label{fig:grid-5x2}
\end{figure}

This procedure gave us, as optimal results, the configuration number 3, with these parameters $L_{ref}=97.86mm$, $t_{ref}=1.33mm$, $H=95.98mm$, $t_\triangle=3.42mm$ and $\alpha=42.68°$, with a value of $PAR=1.37$, $PRR=0.0063$, and as a final value of $J=0.0023$. 

The final 3D model is shown in Fig. \ref{fig:sn_side_by_side}. The polar graphs showing the stiffness and parasitic displacement values while rotating the joint around $\theta_f$ are shown in Fig. \ref{fig:sn_side_by_side}.b, Fig. \ref{fig:sn_side_by_side}.c. The results to obtain these graphs are resumed in Table \ref{tab:stiffness_parerr_45deg} showing the values of stiffness and parasitic error for each values of $\theta_f$. These polar plots provide an immediate visual assessment of how close the joint behavior is to the ideal case. A circular stiffness plot indicates ideal isotropic response. Also the parasitic displacement is shown in the same polar representation to allow a direct visual comparison across $\theta_f$.

To determine which parameters are most sensitive during the feasibility analysis, we individually evaluate the effect of each parameter. In Fig. \ref{fig:grid-5x2} the 2D sweeps one-at-time are shown, in which each input is varied while the other four are held at the best configuration obtained during the feasibility analysis ($L_{ref}=97.86mm$, $H=95.98mm$, $t_{ref}=1.33mm$, $t_\triangle=3.42mm$ and $\alpha=42.68^\circ$). 
\begin{itemize}
    \item \textbf{$L_{ref}$}: IsoErr varies a little ($\approx 0.3- 0.1$) with a plateau after $L_{ref}=40mm$. ParErr increases strongly while moving to a larger length of the panel. The best values of $L_{ref}$ can be found $\approx100mm$.   
    \item\textbf{$H$}: Similar to $L_{ref}$ IsoErr varies slightly between the range $\approx0.32-0.27$ reaching a minimum at low H values. The ParErr grows almost linearly ($\approx5-6$x), reaching its maximum at a higher value of $H$. 
    \item \textbf{$t_{ref}$}: IsoErr increases sharply with $t_{ref}$ ($\approx 6$x) with the minimum at $1.2mm$. ParErr drops markedly ($\approx5$x) as $t_{ref}$ increases. 
    \item \textbf{$t_\triangle$}: The response to IsoErr has a U-shaped trend with net minimum of around $3.5-4 mm$. ParErr Decreases almost monotonically ($\approx4$x) with increasing of the parameter. 
    \item \textbf{$\alpha$}: IsoErr has the highest sensitivity value near $\approx 30^\circ$, with a valley around $\approx 35-40^\circ$. Higher value of $\alpha$ significantly worsens isotropy trend. ParErr has a non-monotonic trend (local peak at $\approx 28-30^\circ$, then increases with the values of $\alpha$.
\end{itemize}

\begin{table}[!t]
\centering
\caption{Stiffness and parasitic error for a 4.5$^\circ$ rotation.}
\label{tab:stiffness_parerr_45deg}
\begin{tabular}{c c c}
\hline
$\theta_f$ [$^\circ$] & Stiffness [N/m] & Parasitic error [mm] \\
\hline
0   & 285.27 & 0.066 \\
10  & 301.02 & 0.092 \\
20  & 318.18 & 0.117 \\
30  & 334.22 & 0.141 \\
40  & 345.61 & 0.159 \\
50  & 349.08 & 0.170 \\
60  & 343.49 & 0.172 \\
70  & 330.63 & 0.164 \\
80  & 314.03 & 0.150 \\
90  & 297.02 & 0.133 \\
100 & 281.86 & 0.113 \\
110 & 269.72 & 0.093 \\
120 & 261.08 & 0.072 \\
130 & 256.06 & 0.052 \\
140 & 254.69 & 0.031 \\
150 & 256.94 & 0.015 \\
160 & 262.83 & 0.021 \\
170 & 272.34 & 0.042 \\
\hline
\multicolumn{3}{l}{\footnotesize \textit{Note} — Values for 180$^\circ$--350$^\circ$ omitted for symmetry.}
\end{tabular}
\end{table}

\subsection{Results Maximum Workspace}

The workspace is one of the key requirements for surgical mechanisms. The joint must provide sufficient rotation for the tip of the endoscope to reach any point inside the patient’s third ventricle. In addition, during the pre-operative planning phase, the EE axis must be aligned with the trajectory selected by the surgeons. The maximum admissible workspace was determined by simulating the joint under fatigue loading until the safety factor reached $SF=1$, resulting in the fatigue-safe rotational workspace $\beta_{ws}$ reported in Table~\ref{tab:workspace_SF1}. For completeness, the table also reports the allowable rotation $\beta_y$ corresponding to the onset of yielding, defined as the rotation at which the maximal von Mises stress reaches the PA12 yield limit ($\sigma_y=45.5~\mathrm{MPa}$).

\begin{table}[!t]
\caption{Workspace $\beta_{ws}$ until $\mathrm{SF}=1$ and allowed rotation $\beta_y$ until reaching the yield stress limit of $\sigma_y=45.5~MPa$}.
\label{tab:workspace_SF1}
\centering
\begin{threeparttable}
\begin{tabular}{ccc}
\hline
$\theta_f$ [$^\circ$] & $\beta_{ws}$ [$^\circ$] & $\beta_y$ [$^\circ$]\\
\hline
0   & 24.37 & 37.75\\
10  & 32.51 & 46.27\\
20  & 34.37 & 47.24\\
30  & 33.25 & 46.26\\
40  & 32.74 & 45.79\\
50  & 28.37 & 42.23\\
60  & 22.69 & 35.93\\
70  & 18.80 & 31.20\\
80  & 16.12 & 27.45\\
90  & 14.26 & 24.66\\
100 & 12.99 & 22.71\\
110 & 12.28 & 21.58\\
120 & 12.12 & 21.37\\
130 & 12.64 & 21.68\\
140 & 13.04 & 22.31\\
150 & 14.23 & 24.10\\
160 & 16.25 & 27.14\\
170 & 19.42 & 31.58\\
\hline
\multicolumn{2}{l}{\footnotesize \textit{Note} — Values for $180^\circ$–$350^\circ$ omitted for symmetry.}
\end{tabular}
\end{threeparttable}
\end{table}

\begin{figure}
    \centering
    \includegraphics[width=0.7\linewidth]{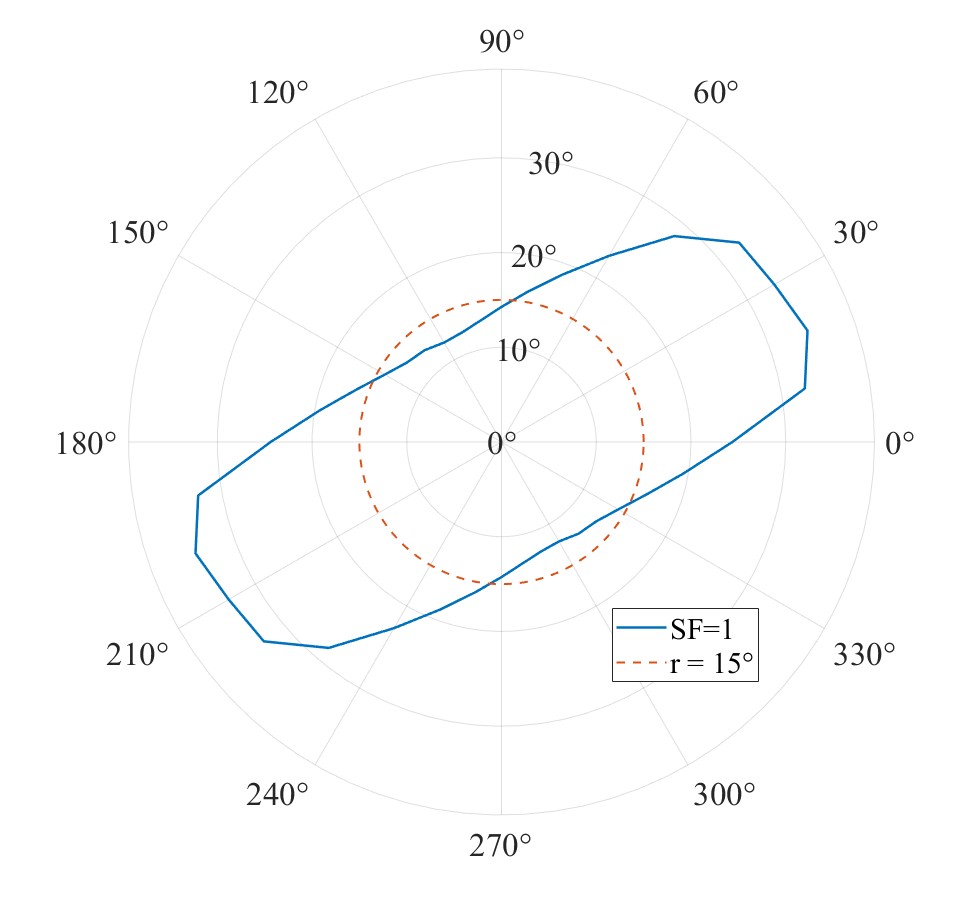}
    \caption{In blue, the rotational workspace of the joint around $\theta_f$. - In red is the minimum workspace for superimposing the EE over the surgical trajectory.}
    \label{Fig.Workspace}
\end{figure}

The polar plot showing the rotational workspace of the joint around the RCM is shown as a blue line in Fig. \ref{Fig.Workspace} with comparison with the minimum workspace, in red, required following the clinical requirements. 

\subsection{Results Validation of the Stiffness Behavior}
Experimental measurements were performed on the SLS-fabricated PA12 prototype following the procedure described in Subsection \ref{validation}. For each of the 12 angular orientations (increments of $30^\circ$ over $360^\circ$ around $\theta_f$), ten displacement samples were acquired under a constant 2 N radial load applied at point N4. The resulting displacements were converted to stiffness values using Eq. (\ref{eq_stiffness}). The stiffness values obtained are shown in Table \ref{tab:exp_sim_comparison} with the Standard Deviations (SD) and the percentage error from the value obtained during the simulation. 


The agreement between experimental and simulated stiffness values was further quantified using MAE, RMSE, MAPE, and mean bias.

\begin{equation}
\mathrm{MAE}=\frac{1}{N}\sum_{i=1}^{N}\left|Exp_i-Sim_i\right|,
\end{equation}

\begin{equation}
\mathrm{RMSE}=\sqrt{\frac{1}{N}\sum_{i=1}^{N}\left(Exp_i-Sim_i\right)^2},
\end{equation}

\begin{equation}
\mathrm{MAPE}=\frac{100}{N}\sum_{i=1}^{N}\frac{\left|Exp_i-Sim_i\right|}{Exp_i},
\end{equation}

\begin{equation}
\mathrm{Bias}=\frac{1}{N}\sum_{i=1}^{N}\left(Exp_i-Sim_i\right),
\end{equation}
where $Exp_i$ is the $i$-th stiffness value obtained experimentally, $Sim_i$ is the corresponding $i$-th stiffness value obtained from the simulation, and $N$ is the number of compared samples.

The resulting global error metrics are reported in Table \ref{tab:error_metrics}.

\begin{table}[h!]
\centering
\caption{Comparison between experimental and simulated stiffness values, with associated percentage error.}
\begin{tabular}{c c c c}
\toprule
& \multicolumn{2}{c}{Stiffness [N/m]} & \\
\cmidrule(lr){2-3}
Angle [$^\circ$] & Exp. mean $\pm$ SD & Sim. & \% Error \\
\midrule
0   & 374.05 $\pm$ 6.14  & 285.27 & 23.73 \\
30  & 479.45 $\pm$ 30.22 & 334.22 & 30.29 \\
60  & 423.41 $\pm$ 17.67 & 343.48 & 18.88 \\
90  & 333.21 $\pm$ 28.82 & 297.02 & 10.86 \\
120 & 289.33 $\pm$ 4.04  & 261.08 & 9.76 \\
150 & 293.16 $\pm$ 10.68 & 256.94 & 12.36 \\
180 & 310.19 $\pm$ 4.41  & 285.27 & 8.03 \\
210 & 315.95 $\pm$ 12.12 & 334.22 & 5.78 \\
240 & 369.67 $\pm$ 5.24  & 343.80 & 7.00 \\
270 & 384.32 $\pm$ 5.71  & 297.02 & 22.72 \\
300 & 355.65 $\pm$ 1.75  & 261.08 & 26.59 \\
330 & 349.69 $\pm$ 7.35  & 256.94 & 26.52 \\
\bottomrule
\end{tabular}
\label{tab:exp_sim_comparison}
\end{table}


\begin{table}[h!]
\centering
\caption{Global error metrics for the comparison between experimental and simulated stiffness values.}
\begin{tabular}{lc}
\toprule
Metric & Value \\
\midrule
MAE [N/m]  & 63.19 \\
RMSE [N/m] & 73.92 \\
MAPE [\%]  & 16.88 \\
Bias [N/m] & 60.15 \\
\bottomrule
\end{tabular}
\label{tab:error_metrics}
\end{table}

\section{Discussion}
We proposed an off-axis compliant RCM joint whose three mobility panels were analytically modified in MATLAB to satisfy an invariant isotropy condition. Conceived as a redesign of Tetra II, the new joint relocates the EE from the center of the tetrahedral structure to its side, improving the surgeon’s line of sight, sterilizability, and the ease of removing the EE in case of malfunction. In addition mobility panels and torsion stiffeners can intersect with each others, allowing a richer set of design solutions while maintaining a small footprint and reducing the risk of self-contact between panels. In simulation, the local anisotropy index approached zero, confirming the isotropy condition governing the EE's directional stiffness. Because achieving isotropy depends on the relative scaling between the panel lengths ($L_i$) and the thicknesses ($t_i$), we adopted the minimum panel length and thickness as reference values ($L_{ref}$ and $t_{ref}$) and the remaining dimensions were obtained from the relations in Table \ref{tab:isotropic_params}. A second verification was performed by simulating the three mobility panels by varying $L_{ref}$ and $t_{ref}$ in three scenarios, obtaining as ratios between the semi axes $PAR_1=1.12$, $PAR_2=1.06$ and $PAR_3=1.10$. These results, close to unit values, confirm that geometry ratios, rather than absolute dimensions, govern the EE’s directional stiffness.

The constraining panels, which maintain the RCM, were added via a feasibility analysis using ANSYS Workbench 2023 R1. Their dimensions were guided by two scalar metrics: IsoErr, which quantifies isotropy (lower is better), and ParErr, which measures the ratio of EE motion to RCM drift (higher is better). 

To determine which parameters are most sensitive during the feasibility analysis, we individually evaluate the effect of each parameter in Fig. \ref{fig:grid-5x2}. Specifically for $L_{ref}$ and $H$ both parameters exhibit weak coupling to IsoErr across the explored ranges and strong, favorable trends in ParErr. Practically, this means they are the safest knobs to improve parasitic behavior late in the design, provided packaging, mass, and draping constraints are respected. The observed “best” values lie near the upper end of the tested ranges, suggesting additional headroom might exist if the envelope permits, but extrapolation beyond the sweep should be validated with updated constraints. The parameter $t_{ref}$ is one the most sensitive. In fact increasing the baseline panel thickness sharply worsens isotropy and simultaneously reduces ParErr in this design space. In short, thicker "reference” panels make both metrics worse. As a result, $t_{ref}$ should be kept as low as manufacturability and strength allow, and not used as a tuning knob for parasitics. The parameter $t_\triangle$ shows a U‑shaped IsoErr with a clear minimum at a moderate thickness. Meanwhile, ParErr declines as $t_\triangle$ grows. The implication is a sweet spot: set $t_\triangle$ near the isotropy minimum and only deviate if you must trade some isotropy for parasitic behavior. The angle $\alpha$ is another sensitive parameter, in fact there is a narrow valley that preserves near‑isotropic stiffness, while small departures degrade it quickly. 
Overall, the sweeps recommend a tuning workflow: lock isotropy with $\alpha$ and $t_\triangle$, then raise ParErr with $H$ and $L_{ref}$, while keeping $t_{ref}$ minimal.

When assembled with the constraining panels, the joint retained near-isotropic behavior (ellipse PAR up to 1.37) while preserving the RCM (PRR $=0.0063$), obtaining as normalized value of $J=0.0023$ and thus a reduction in parasitic error with quasi-isotropic stiffness. Further FEM analyses revealed that across $0$–$360^\circ$ actuation, parasitic RCM drift under a $4.5^\circ$ command remained sub-millimetric (0.015–0.172~mm), comfortably within the 1 mm clinical threshold and therefore unlikely to increase the risks of tissue injury significantly. Analyzing the polar plot of Fig. \ref{fig:sn_side_by_side}.b and \ref{fig:sn_side_by_side}.c the highest parasitic displacement values occur near the angular regions where the stiffness polar plot deviates most from a circle. This observation is consistent with the statement made at the Uniform stiffness behavior requirement introduced in Section \ref{Requirements}, and further motivates the pursuit of near-isotropic stiffness to limit parasitic RCM shift.

The fatigue-bounded workspace ($SF=1$) spanned $12.1^\circ$–$34.4^\circ$ depending on orientation. Although most directions exceeded the $\pm 15^\circ$ clinical target, a subset ($\approx 90^\circ$–$140^\circ$) fell modestly short ($\approx 12^\circ$–$14^\circ$), indicating a trade-off between uniform workspace and conservative fatigue safety. This suggests trajectory planning that avoids the least compliant orientations.

\begin{table*}[!t]
\caption{Comparison between compliant RCM joint variants and their main performance indicators.}
\label{tab:joint_comparison}
\centering

\renewcommand{\arraystretch}{1.25}
\setlength{\extrarowheight}{1pt}

\begin{threeparttable}
\begin{tabular}{
  >{\raggedright\arraybackslash}p{0.23\linewidth}
  >{\raggedright\arraybackslash}p{0.44\linewidth}
  >{\raggedright\arraybackslash}p{0.30\linewidth}
}
\hline
\textbf{\mbox{Mechanism}} & \textbf{Description} & \textbf{Key features} \\
\hline
\textbf{Tetra II \cite{rommers2021new}} &
A compliant RCM joint based on a tetrahedral flexure architecture, which maintains a fixed remote center of motion while steering the EE. &
The EE is located on the central axis of the tetrahedral structure, which limits surgical usability in the operating room. \\
\hline
\textbf{Off-axis compliant mechanism (unoptimized) \cite{mariano2026dual}} &
The EE holder is shifted from the center of the tetrahedral structure to a lateral (off-axis) position to improve surgical accessibility. &
Stiffness is highly anisotropic ($PAR = 21.63$) and the peak parasitic error is relatively high ($1.10~mm$). \\
\hline
\textbf{Dual-joint compliant mechanism \cite{mariano2026dual}} &
Two identical off-axis joints are connected in series to reduce parasitic RCM displacement and improve stiffness isotropy. &
Stiffness anisotropy is reduced (PAR = 1.55) with a lower peak parasitic error (0.051~mm). \\
\hline
\textbf{Off-axis compliant joint (optimized)} &
Starting from the Tetra II architecture, the geometry is optimized, via feasibility analysis, to reduce parasitic motion while maintaining near-isotropic stiffness. &
Near-isotropic stiffness (PAR = 1.37) with minimal peak parasitic error (0.172~mm). \\
\hline
\end{tabular}

\end{threeparttable}
\end{table*}

Bench measurements with a 2~N load reproduced the trend of directional stiffness, with a simulation error typically 6–30\%. The largest discrepancies clustered in $30^\circ$ ($\approx 30.3\%$), $300^\circ$ ($\approx 26.6\%$), $330^\circ$ ($\approx 26.5\%$), $0^\circ$ ($\approx 23.7\%$) and $270^\circ$ ($\approx 22.7\%$), while the smallest appeared at $210^\circ$ ($\approx 5.8\%$), $240^\circ$ ($\approx 7.0\%$) and $180^\circ$ ($\approx 8.0\%$), with still low values at $120^\circ$ ($\approx 9.8\%$) and $90^\circ$ ($\approx 10.9\%$). These variations can be explained by differences between the built joint and the nominal CAD/FEM geometry. These difference may depends by deviations and local imperfections introduced during SLS printing of PA12, like minor thickness, geometry variations and micro-defects, which can affect the effective stiffness distribution and slightly distort the fitted stiffness ellipse with respect to the FEM prediction. As a result, the local difference between the experimental and simulated stiffness trends can become more pronounced at some orientations. A possible way to reduce these discrepancies is to improve the repeatability of the built part. This can be obtained either by adopting alternative powder-bed processes such as Multi Jet Fusion (MJF), which has been reported to yield PA12 parts with slightly lower porosity and a more homogeneous mechanical response compared to SLS \cite{rosso2020depth}, or by introducing a calibration loop: critical dimensions like the flexure-wall thickness, are measured on the printed prototype (e.g., with a caliper) and the CAD model is then updated accordingly until the nominal design matches the measured geometry.

A more global interpretation of the experimental--numerical agreement can be obtained from the error metrics reported in Table \ref{tab:error_metrics}. The MAE of $63.19~N/m$ and the RMSE of $73.92~N/m$ indicate a moderate discrepancy between the simulated and measured stiffness values over the 12 tested orientations. Since the RMSE is only moderately larger than the MAE, the mismatch does not appear to be dominated by a small number of extreme outliers, but rather to be distributed across multiple directions. The MAPE of $16.88\%$ suggests that the FEM model provides a reasonable estimate of the directional stiffness behavior, while still leaving room for improvement in the prediction of the absolute stiffness magnitude. Finally, the positive mean bias of $60.15~N/m$ indicates that the simulations tend to systematically underestimate the experimental stiffness.

The experiments were performed using a dead weight of $204\pm1~\mathrm{g}$, connected to the joint through a cable and a pulley. This setup may introduce several sources of error, including pulley friction, string compliance, micro-slip between the cable and the pulley, and measurement noise. To estimate the effective force transmitted to the joint during the experiments, we replaced the joint with an ATI Nano17 F/T sensor and measured the force along the loading direction. The measured mean force was $1.978~\mathrm{N}$, with a maximum deviation of $+0.0104~\mathrm{N}$ and a minimum deviation of $-0.0078~\mathrm{N}$ around the mean. Comparing the expected applied force ($1.996$--$2.006~\mathrm{N}$) with the measured transmitted force ($1.9702$--$1.9884~\mathrm{N}$), the resulting force transmission error due to parasitic effects is bounded between $0.38\%$ and $1.78\%$, with a nominal mean error of approximately $1.15\%$.

Beyond mechanical performance, the new configuration provides practical instrument management, visualization, and sterility advantages. First, because the EE is now external to the joint structure rather than housed within it, it can be safely and rapidly disengaged without withdrawing it from its working cavity/portal. This enables faster instrument exchange and immediate manual override in the event of a malfunction, reducing intraoperative downtime and risk. Second, visibility improves because there is less joint volume around the EE axis, minimizing occlusion near the entry portal and preserving the line of sight to the surgical field. Third, sterilizability is enhanced by the clearer physical separation between the internal structure (which 
 does not need to be sterilized and can remain behind a sterile barrier) and the external structure (which is intended for sterilization). This partition simplifies sterile draping, eases workflow, and supports robust sterilization methods for patient-facing components.

Compared with double-parallelogram or spherical RCM wrists, the proposed monolithic compliant joint avoids the use of discrete joints and bearings while intrinsically enforcing an RCM. Compared to software-RCM approaches, the geometry itself constrains motion, simplifies control, improves fail-safe behavior, and, in this new configuration, streamlines EE release, enhances visibility, and clarifies the sterilization pathway.

Finally, Table~\ref{tab:joint_comparison} compares the optimized off-axis compliant joint with other tetrahedral RCM compliant mechanisms, highlighting their key features and performance metrics.

\section{Conclusion}
We introduced an off-axis monolithic compliant RCM joint whose three mobility panels were analytically shaped to satisfy an invariant isotropy condition. We verified, via ANSYS and bench tests, that geometry ratios, rather than absolute dimensions, govern directional stiffness at the EE. With the addition of constraining panels selected using a merit metric, the joint retains quasi-isotropic behavior with values of PAR $\leq 1.37$, PRR $= 0.0063$ and $J = 0.0023$. PAR quantifies stiffness anisotropy as the ratio between the major and minor semiaxes of the fitted stiffness ellipse, PRR measures the fraction of commanded EE rotation lost to RCM drift and J is a normalized index use for indicating the best results. Lower values of J indicate the best trade off between PAR and PRR. Across $0$–$360^\circ$ actuation, a $4.5^\circ$ command yielded submillimetric RCM drift (0.015–0.172 mm), within the clinical threshold of 1 mm. The fatigue-bounded workspace (SF $=1$) spans $12.1^\circ$–$34.4^\circ$, exceeding the clinical target of $\pm 15^\circ$ in most directions, with a modest shortfall around $90^\circ$–$140^\circ$ ($\approx 12^\circ$–$14^\circ$). Beyond mechanics, the new configuration offers practical advantages: placing the EE external to the joint enables safe, rapid detachment without withdrawing it from its cavity, reducing joint volume around the EE axis improves line-of-sight, and a clean separation between internal (non-sterilizable) and external (sterilizable) structures simplifies the sterile workflow.

Future work will investigate materials with higher fatigue strength, such as titanium alloys and NiTi, to address the insufficient workspace observed in specific orientations. In particular, austenitic (superelastic) nitinol can withstand millions of cycles without fracture, allowing longer fatigue-limited angular excursions \cite{mahtabi2015fatigue}. Its biocompatibility further supports intraoperative use \cite{wadood2016brief}. Moreover, recent studies show that superelastic NiTi can be additively manufactured via 3D printing, enabling the fabrication of complex application-specific components \cite{yan2023superelastic, yan2024correlation, lantada2024additive}.

In parallel, we plan to implement a ball screw-based manual actuation to provide controlled motion of the EE, replacing the EE with an endoscope for intraoperative repositioning, while preserving the RCM constraint. We will also investigate compact motorization to realize a fully robotic version with closed-loop control and trajectory planning. Finally, accuracy and usability will be evaluated on a head phantom, benchmarking tip precision and setup time against conventional multi-component wrists \cite{mariano2024design}.

\section*{Declaration of competing interest}
The authors declare that they have no known competing financial interests or personal relationships that could have appeared to influence the work reported in this paper.

\ifCLASSOPTIONcaptionsoff
  \newpage
\fi



%
\bibliographystyle{IEEEtran}

\bibliography{IEEEexample.bib}

%








\end{document}